\documentclass[letterpaper, 10 pt, conference]{ieeeconf}  

\IEEEoverridecommandlockouts                              

\usepackage{float}
\usepackage{amsmath}
\usepackage{multirow}
\usepackage[ruled,linesnumbered]{algorithm2e}
\usepackage{amssymb}
\usepackage{comment}
\usepackage{caption}
\usepackage{cite}

\usepackage{mathtools}
\usepackage{subcaption}
\usepackage{booktabs}
\usepackage{color}
\usepackage{siunitx}
\usepackage{algorithmic}
\usepackage[ruled, linesnumbered]{algorithm2e}

\usepackage[whole]{bxcjkjatype}

\newcommand{\argmin}{\mathop{\mathrm{arg~min}}\limits}

\newcommand{\figref}[1]{Fig.~\ref{#1}}                          
\newcommand{\tabref}[1]{Table~\ref{#1}}

\title{\LARGE \bf
Task-Relevant and Irrelevant Region-Aware Augmentation \\ for Generalizable Vision-Based Imitation Learning \\ in Agricultural Manipulation
}

\author{Shun Hattori$^{1}$, Hikaru Sasaki$^{1}$, Takumi Hachimine$^{1}$, Yusuke Mizutani$^{2}$, and Takamitsu Matsubara$^{1}$
\thanks{$^{1}$S. Hattori, H. Sasaki, T. Hachimine, and T. Matsubara are with the Division of Information Science, Graduate School of Information Science, Nara Institute of Science and Technology (NAIST), Nara, Japan. email: {\tt\small hattori.shun.ho5@naist.ac.jp, hachimine.takumi.hs8@is.naist.jp, sasaki.hikaru@is.naist.jp, takam-m@is.naist.jp}}
\thanks{$^{2}$Y. Mizutani is with DX \(\cdot\) IT \(\cdot\) Research \& Development Center, TSUBAKIMOTO CHAIN CO., Kyoto, Japan. email: {\tt\small y.mizutani@gr.tsubakimoto.co.jp}}
}

\begin{document}

\maketitle
\thispagestyle{empty}
\pagestyle{empty}


\begin{abstract}
Vision-based imitation learning has shown promise for robotic manipulation; however, its generalization remains limited in practical agricultural tasks.
This limitation stems from scarce demonstration data and substantial visual domain gaps caused by i) crop-specific appearance diversity and ii) background variations.
To address this limitation, we propose Dual-Region Augmentation for Imitation Learning (DRAIL), a region-aware augmentation framework designed for generalizable vision-based imitation learning in agricultural manipulation.
DRAIL explicitly separates visual observations into task-relevant and task-irrelevant regions.
The task-relevant region is augmented in a domain-knowledge-driven manner to preserve essential visual characteristics, while the task-irrelevant region is aggressively randomized to suppress spurious background correlations.
By jointly handling both sources of visual variation, DRAIL promotes learning policies that rely on task-essential features rather than incidental visual cues.
We evaluate DRAIL on diffusion policy-based visuomotor controllers through robot experiments on artificial vegetable harvesting and real lettuce defective leaf picking preparation tasks.
The results show consistent improvements in success rates under unseen visual conditions compared to baseline methods.
Further attention analysis and representation generalization metrics indicate that the learned policies rely more on task-essential visual features, resulting in enhanced robustness and generalization.
\end{abstract}

\section{Introduction}
Vision-based imitation learning~\cite{chi2023DP, chi2024universal} enables robotic manipulation without explicit feature engineering, yet achieving robust generalization in real-world agricultural environments remains challenging~\cite{kim2025Pepper}.
A primary challenge is the high cost of collecting real-robot demonstrations in agricultural environments.
Unlike structured industrial settings, agricultural scenes exhibit substantial variability in crop shapes, growth stages, lighting conditions, and occlusions, and are often constrained by seasonal and environmental factors.
These characteristics make large-scale data collection time-consuming and costly, requiring policies to generalize effectively from limited demonstrations.

\begin{figure}[t]
    \centering
    \begin{minipage}[b]{1.0\linewidth}
        \centering
        \includegraphics[width=0.95\linewidth]{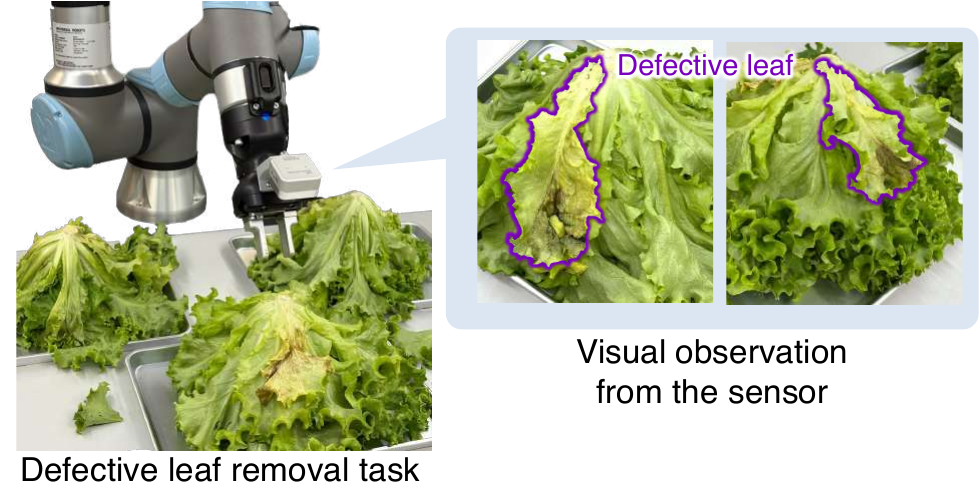}
        \subcaption{Robot automation system}
        \label{fig:overview:task}
    \end{minipage}
    \begin{minipage}[b]{1.0\linewidth}
        \centering
        \includegraphics[width=0.95\linewidth]{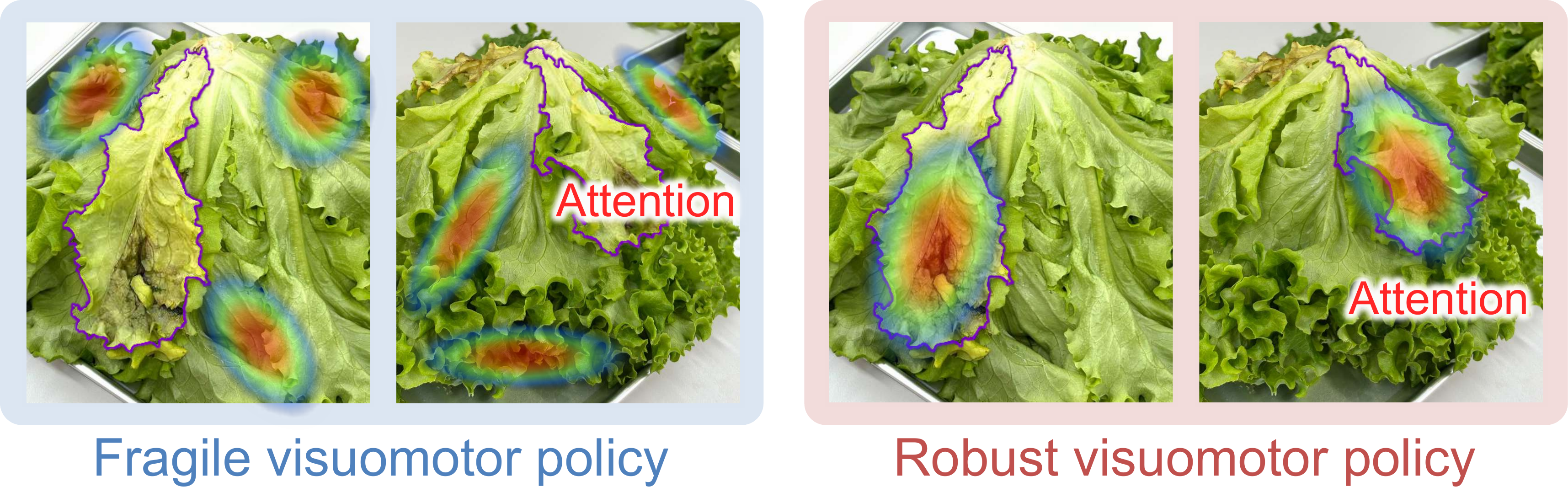}
        \subcaption{Attention regions of visuomotor policy}
        \label{fig:overview:attention}
    \end{minipage}
    \caption{
    Overview of an agricultural manipulation and the importance of visual attention of visuomotor policies.
    (a) Robot automation system for a defective leaf removal task, the visuomotor policy must select appropriate actions from visual observations.
    (b) Fragile visuomotor policies tend to attend to various regions in the visual observation, whereas robust policies focus their attention on the defective leaf.
    }
    \label{fig:overview}
\end{figure}

Under such data-scarce conditions, visual variability becomes particularly problematic.
In real-world agricultural environments, visual inputs exhibit two major sources of variation: i) crop-specific appearance diversity and ii) background variations.
These factors enlarge the domain gap between demonstration and deployment, degrading policy performance under unseen visual conditions.
This limitation arises because visual observations contain not only task-relevant information but also substantial task-irrelevant and redundant cues.
When demonstration data are scarce, policies tend to overfit spurious background correlations rather than task-essential visual features, leading to poor generalization (\figref{fig:overview}).

Prior works, not limited to agricultural domains, have introduced domain randomization and data augmentation techniques to diversify visual observations~\cite{yuRSS2023, chenIJRR2025, ishida2024TaskRel}.
Some methods leverage image generation models to alter object appearance and backgrounds while preserving task-relevant information~\cite{yuRSS2023, chenIJRR2025}.
Others use segmentation-based approaches to randomize backgrounds and reduce overfitting to specific environmental cues~\cite{ishida2024TaskRel}.
However, existing approaches typically address either background diversity or overall appearance diversity, and often rely on large amounts of training data to compensate for the remaining source of variation.
Importantly, these two sources of variation coexist in agricultural environments, making it insufficient to address them independently.

\begin{figure*}[t]
    \centering
    \includegraphics[width=0.95\linewidth]{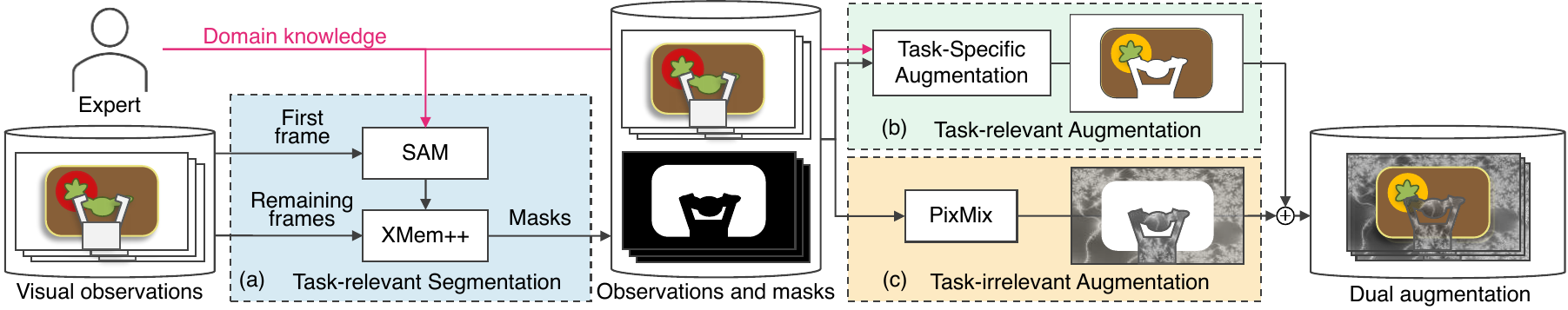}
    \caption{
    Overview of Dual-Region Augmentation for Imitation Learning (DRAIL).
    (a) shows task-relevant segmentation.
    A mask of the task-relevant region is initialized on the first frame of visual observations using SAM and then propagated to subsequent frames using XMem++, yielding per-frame masks.
    (b) shows task-relevant augmentation.
    The task-relevant region undergoes task-specific augmentation.
    (c) shows task-irrelevant randomization.
    The task-irrelevant region is perturbed using PixMix.
    The two streams are composited to form augmented data, repeated across all demonstration episodes.
    }
    \label{fig:drail_diagram}
\end{figure*}

In this paper, we propose Dual-Region Augmentation for Imitation Learning (DRAIL), a region-aware augmentation framework designed for generalizable vision-based imitation learning in agricultural manipulation.
DRAIL explicitly separates visual observations into task-relevant and task-irrelevant regions.
The task-relevant region is augmented in a domain-knowledge-driven manner to preserve essential visual characteristics, while the task-irrelevant region is aggressively randomized to suppress spurious background correlations.
By jointly handling both sources of visual variation, DRAIL promotes learning policies that rely on task-essential features rather than incidental visual cues.

We validate DRAIL on diffusion policy-based visuomotor controllers~\cite{chi2023DP} through robot experiments on artificial vegetable harvesting and real lettuce defective leaf picking preparation tasks.
Experimental results demonstrate improved success rates under unseen visual conditions compared to prior methods. Attention analysis and representation generalization metrics further confirm that DRAIL encourages policies to focus on task-essential visual features, leading to enhanced robustness and generalization.

Our contributions are threefold:
\begin{itemize}
    \item We propose DRAIL, a dual-region augmentation framework for generalizable vision-based imitation learning.
    \item We provide empirical design examples of task-relevant region augmentation across multiple agricultural manipulation tasks.
    \item We demonstrate that DRAIL improves generalization under unseen visual conditions on robot agricultural manipulation tasks using diffusion policy–based controllers.
\end{itemize}

\section{Related Works}
\subsection{Automation of Agricultural Tasks with Imitation Learning}
Kim et al.~\cite{kim2025Pepper} proposed a pepper harvesting system using a mobile manipulator and vision-based imitation learning.
They collected demonstration data using a demonstration device~\cite{chi2024universal} and learned a policy based on a diffusion model~\cite{chi2023DP}.
In their experiment, the task success rate was limited to $28.95\%$.
The authors attributed the failures primarily to the visuomotor policy overfitting to task-irrelevant visual features. As a result, the policy generated motions based on spurious visual correlations, leading to inaccurate alignment. This suggests that visuomotor policies struggle to consistently identify task-essential visual features under visual variability, and that changes in object appearance and background can directly cause task failure.

In contrast, while our method also adopts vision-based imitation learning for agricultural tasks, we introduce task-relevant and task-irrelevant region-aware augmentation driven by domain knowledge. By explicitly separating visual observations into task-relevant and task-irrelevant regions and applying differentiated augmentation strategies, DRAIL suppresses spurious background correlations and promotes reliance on task-essential visual features. Consequently, the learned visuomotor policy achieves improved generalization under unseen visual conditions.

\subsection{Data Augmentation in Imitation Learning}
Augmenting the demonstration dataset in IL has been used to prevent a visuomotor policy from depending on limited visual features in demonstration, thereby improving generalization performance to unseen environments.
Prior work has proposed randomizing textures, illumination, and backgrounds in simulation~\cite{James2017transf}, randomizing backgrounds with fractal images after segmenting backgrounds using a foundation model~\cite{ishida2024TaskRel}, and compositing task-aligned backgrounds using diffusion models~\cite{yuan2025roboengine}.
These approaches share the common idea of introducing appearance changes during learning so that the policy is encouraged to act based on task-essential visual information.

Unlike prior approaches that primarily randomize backgrounds, our method additionally augments task-relevant regions using domain knowledge. This design facilitates data augmentation under limited demonstrations and encourages the policy to rely more on task-essential visual features, thereby mitigating overfitting to spurious correlations.

\section{Dual-Region Augmentation for Imitation Learning}
In this section, we propose Dual-Region Augmentation for Imitation Learning (DRAIL), shown in \figref{fig:drail_diagram}, a framework that achieves visual domain generalization in IL for agricultural tasks via region-aware data augmentation.
DRAIL combines: i) task-specific data augmentation guided by domain knowledge applied to task-relevant regions, and ii) strong randomization of task-irrelevant regions to discourage reliance on spurious cues.
By jointly handling these two aspects, DRAIL reduces over-reliance on irrelevant information and improves robustness to out-of-distribution (OOD) variations in task-relevant regions.

\subsection{Imitation Learning with Data Augmentation}
In vision-based imitation learning, we learn a visuomotor policy from demonstration data
$\mathcal{D}=\{(\mathbf{o}_i,\mathbf{s}_i,\mathbf{a}_i)\}_{i=1}^{N}$,
which consists of $N$ tuples of visual observation, state, and action.
The visuomotor policy predicts the action $\mathbf{a}$ from the visual observation $\mathbf{o}$ and the state $\mathbf{s}$, and is written as
$\pi_\theta(E_\phi(\mathbf{o}),\mathbf{s})$,
where, $\theta$ denotes the parameters of the visuomotor policy, and $E_\phi$ is an image encoder parameterized by $\phi$.

We introduce visual augmentation $f_\mathrm{aug}$ as a pre-processing for the demonstration data.
The augmented demonstration data $\mathcal{D}_\mathrm{aug}$ is obtained by applying the augmentation $f_\mathrm{aug}$ to the visual observations in the demonstration data $\mathcal{D}$.
The parameters of the visuomotor policy and the image encoder,
$\theta^\ast,\; \phi^\ast$, are optimized by minimizing the prediction error:
\begin{align}
    \theta^\ast, \phi^\ast =
    \argmin_{\theta, \phi} ~ \sum_{\mathbf s, \mathbf a, \tilde{\mathbf o}\in \mathcal{D}_\mathrm{aug}} \left\| \mathbf a - \pi_\theta(E_\phi(\tilde{\mathbf{o}}),\mathbf{s}) \right\|_2^2,
    \label{eq:policy_opt}
\end{align}
where $\tilde{\mathbf{o}}_i$ is the augmented visual observation, given by $\tilde{\mathbf{o}}_i = f_\mathrm{aug}(\mathbf o_i)$.

\subsection{Visual Observation Augmentation}
We augment the visual observations $\mathbf{o}$ in the demonstration data $\mathcal D$ by independently augmenting the task-relevant region and the task-irrelevant region.
The task-relevant region contains visual information required for the visuomotor policy to infer actions to accomplish the task.
The task-irrelevant region is any part of the visual observation that is not task-relevant.

Let $\mathbf{o}\in\mathbb{R}^{H\times W\times C}$ be a visual observation and
$M\in\left\{0, 1\right\}^{H \times W}$ be the task-relevant region mask.
Let $p_{\mathrm{rel}}$ denote the task-specific augmentation applied to the task-relevant region, and
$p_{\mathrm{irr}}$ denotes the strong randomization applied to the task-irrelevant region.
The augmented visual observation $\tilde{\mathbf{o}}$ is then computed as:
\begin{align}
    \tilde{\mathbf{o}} =
    p_{\mathrm{rel}}( \mathbf{o} \odot M) \oplus p_{\mathrm{irr}}(\mathbf{o} \odot (\mathbf{1}-M)),
\end{align}
where $\odot$ denotes Hadamard product, and $\oplus$ denotes element-wise compositing.

\subsubsection{Task-Relevant Region Extraction}
Task-relevant region extraction corresponds to \figref{fig:drail_diagram} (a).
Using a segmentation foundation model and video object segmentation (VOS), we create the task-relevant region mask $M$.
In our implementation, we use Segment Anything Model (SAM)~\cite{Kirillov2023SAM} as the segmentation foundation model and XMem++~\cite{Bekuzarov2023XMEM2} for VOS.
By annotating the task-relevant region on the initial frame of each demonstration, we can automatically propagate masks and generate masks for subsequent visual observations.

\subsubsection{Task-Relevant Region Augmentation}
Task-relevant region augmentation corresponds to \figref{fig:drail_diagram} (b).
This process is designed to improve the generalization of the visual observation beyond the limited appearance in the demonstration data and, thus, must be tailored to the task using domain knowledge.
While preserving task-essential image features, we add image features for which the same action is expected.

\subsubsection{Task-Irrelevant Region Augmentation}
Task-irrelevant region augmentation corresponds to \figref{fig:drail_diagram} (c).
Following Ishida et al.~\cite{ishida2024TaskRel}, we apply an overlay of fractal textures with high geometric complexity to the task-irrelevant region.
This fractal-texture-based randomization overlays a large variety of fractal textures onto images, and we use PixMix~\cite{Hendrycks2022PixMix} for implementation.

\section{Experiment}
We investigate whether the proposed method, DRAIL, can learn a visuomotor policy that generalizes to unseen visual variations in crop shape and color that are not included in the demonstration dataset.
The research questions addressed in our experiments are as follows:
\begin{enumerate}
    \item[RQ1:] Do the dual-region augmentation components of DRAIL improve generalization performance of learned policies under unseen visual variations?
    \item[RQ2:] Does DRAIL promote attention to task-essential visual features while suppressing spurious background correlations?
    \item[RQ3:] 
    Can the generalization ability of DRAIL be quantitatively evaluated?
\end{enumerate}
We evaluate RQ1–RQ3 independently for each of the three tasks.

\subsection{General Settings}
As comparison methods, we use DRAIL without task-relevant augmentation~\cite{ishida2024TaskRel}, DRAIL without task-irrelevant augmentation, and DRAIL without dual augmentation~\cite{chi2023DP}.
For all methods, we use diffusion policy~\cite{chi2023DP} as the visuomotor policy model, with ResNet~\cite{hekaiming2016resnet} as the image encoder and UNet~\cite{ronneberger2015unet} as the denoising network.

To evaluate DRAIL and three ablation methods, we adopt three metrics corresponding to RQ1-RQ3:
(i) task success rates, including motion selection and positional alignment, to assess policy performance (RQ1);
(ii) visualization of policy attention using saliency maps~\cite{simonyan2014saliency}, to analyze whether the policy focuses on task-relevant regions (RQ2); and
(iii) visual generalization quantified by Absolute RND Gap (ARG), defined as the absolute difference between RND values~\cite{burda2019RND} computed using demonstration and test data, where smaller values indicate better generalization (RQ3).

\begin{figure}[t]
    \centering
    \begin{minipage}[b]{0.7\linewidth}
        \centering
        \includegraphics[width=0.95\linewidth]{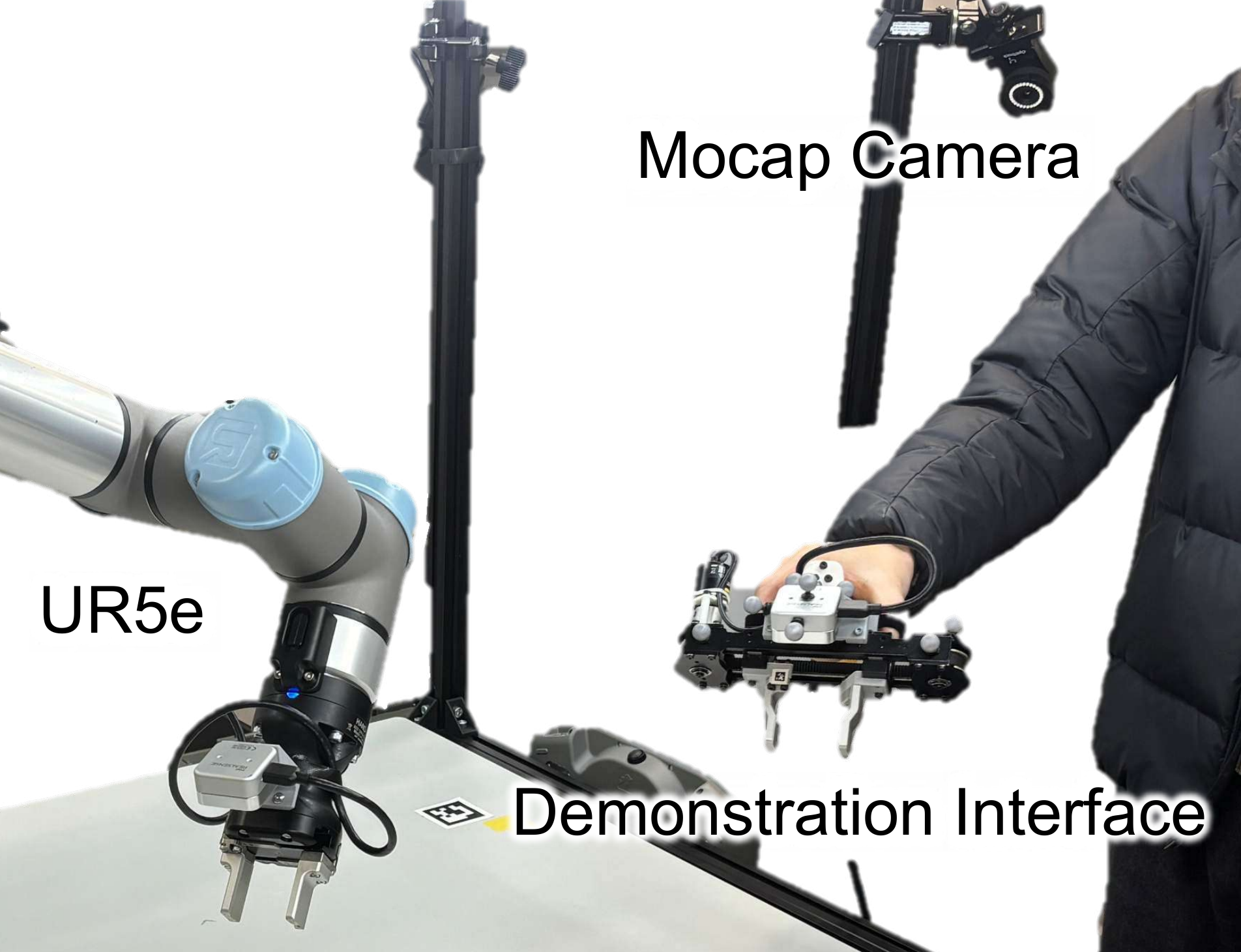}
        \subcaption{Robot system}
        \label{fig:operation_system:gripper}
    \end{minipage}
    \begin{minipage}[b]{0.28\linewidth}
        \begin{minipage}[b]{1.0\linewidth}
            \centering
            \includegraphics[width=1.0\linewidth]{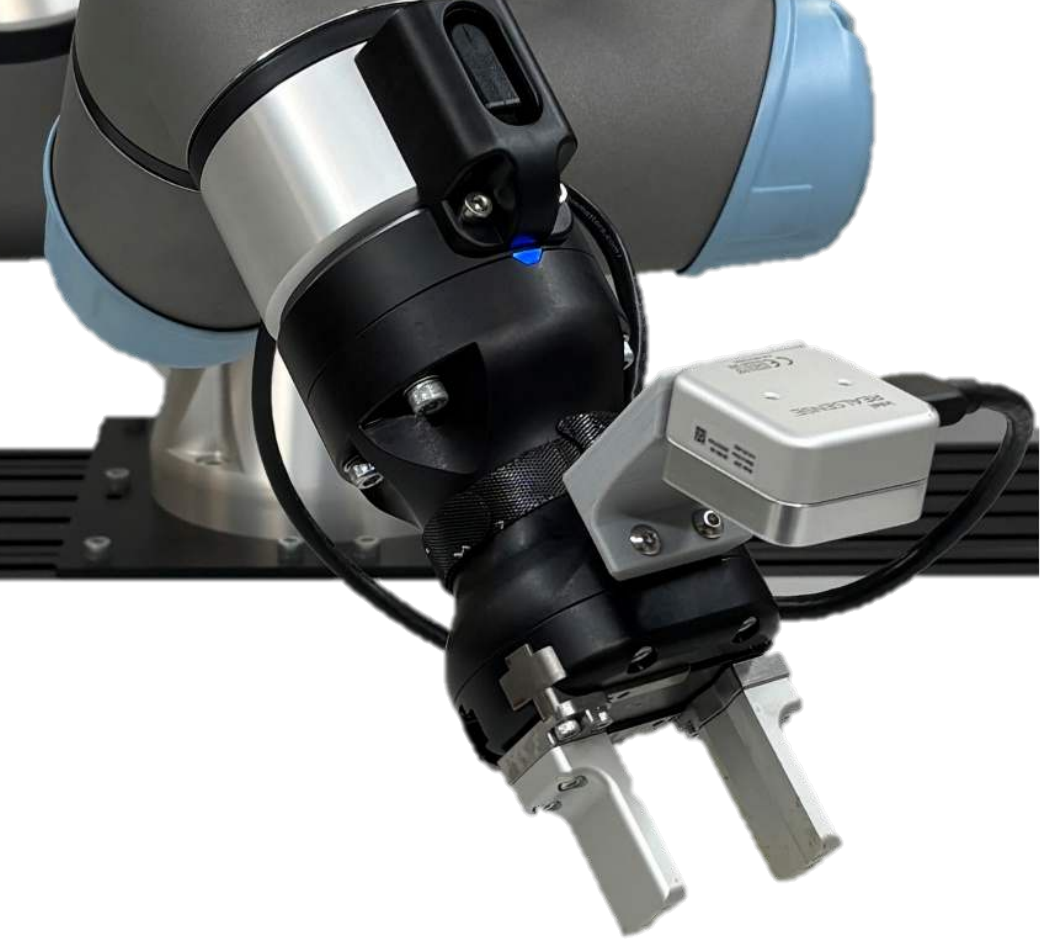}
            \subcaption{Robot gripper}
            \label{fig:operation_system:gripper}
        \end{minipage}
        \begin{minipage}[b]{1.0\linewidth}
            \centering
            \includegraphics[width=1.0\linewidth]{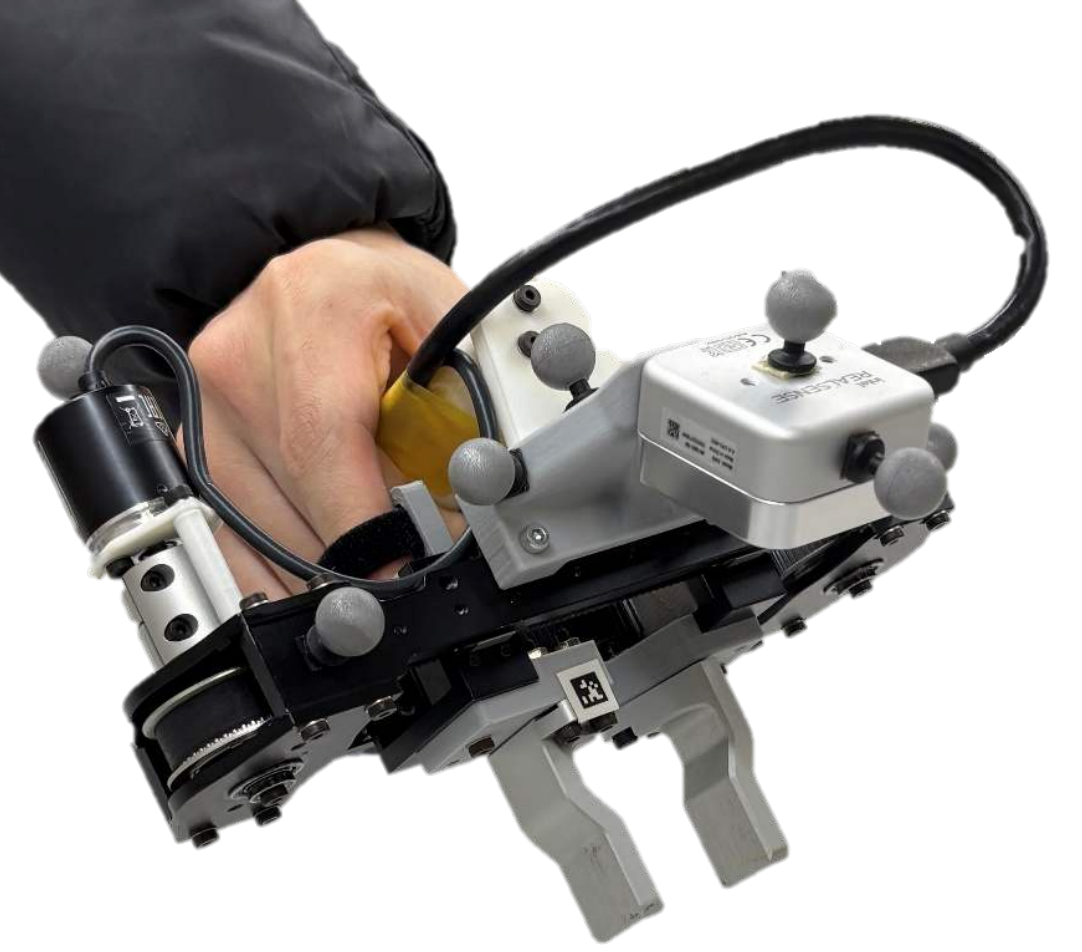}
            \subcaption{Demo interface}
            \label{fig:operation_system:interface}
        \end{minipage}
    \end{minipage}
    \caption{Robot hardware used in experiments.}
    \label{fig:operation_system}
\end{figure}

\figref{fig:operation_system} shows the robot system used in our robot experiments.
For demonstration, we use a two-finger opposing demonstration interface that has the same end-effector structure as the robot gripper.
The interface is equipped with motion-capture markers to measure position and orientation, as well as an encoder to measure the gripper opening.
During the demonstration, the measured pose and opening are synchronized with the robot.

The state $\mathbf{s}_t$ is the end-effector pose of the robot, $\mathbf{o}_t$ is the visual observation captured by a camera mounted on the robot gripper, and the action $\mathbf{a}_t$ is the target pose of the robot.
The control frequency of the robot system is 10~\si{\hertz}.

\begin{figure}[t]
    \centering
    \begin{minipage}[b]{0.49\linewidth}
        \centering
        \includegraphics[width=0.8\linewidth]{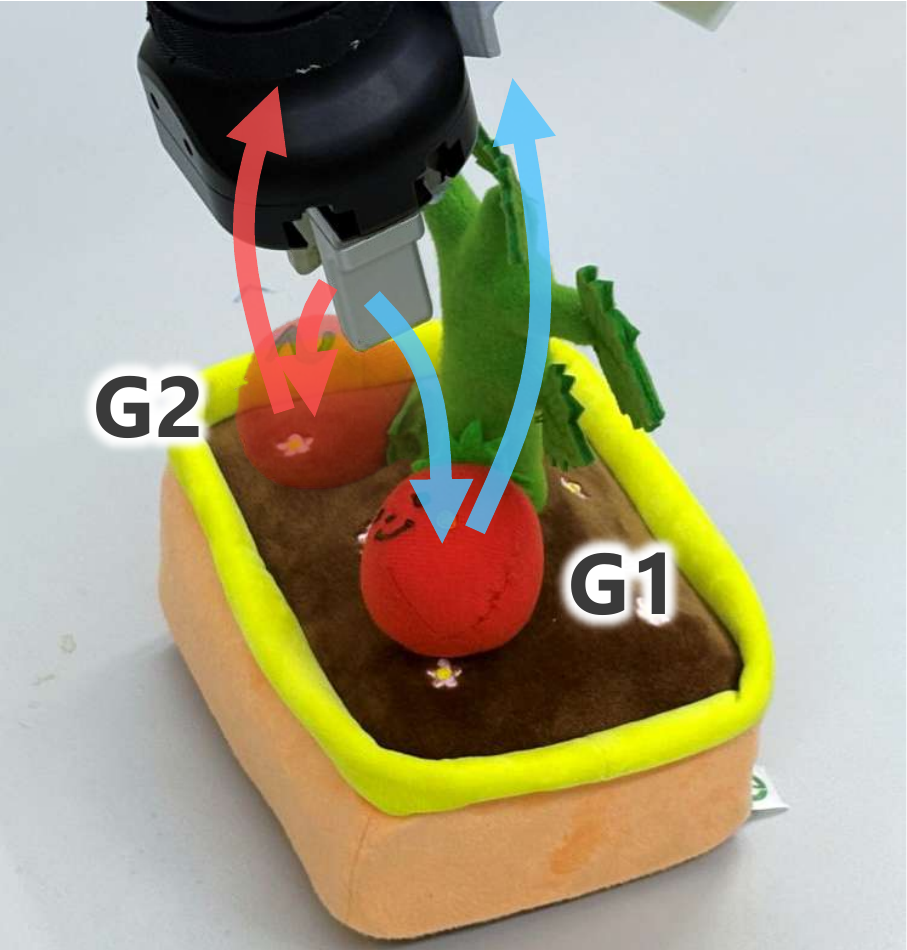}
        \subcaption{Demonstration environment}
        \label{fig:tomato_task_env:demo}
    \end{minipage}
    \begin{minipage}[b]{0.49\linewidth}
        \centering
        \includegraphics[width=0.8\linewidth]{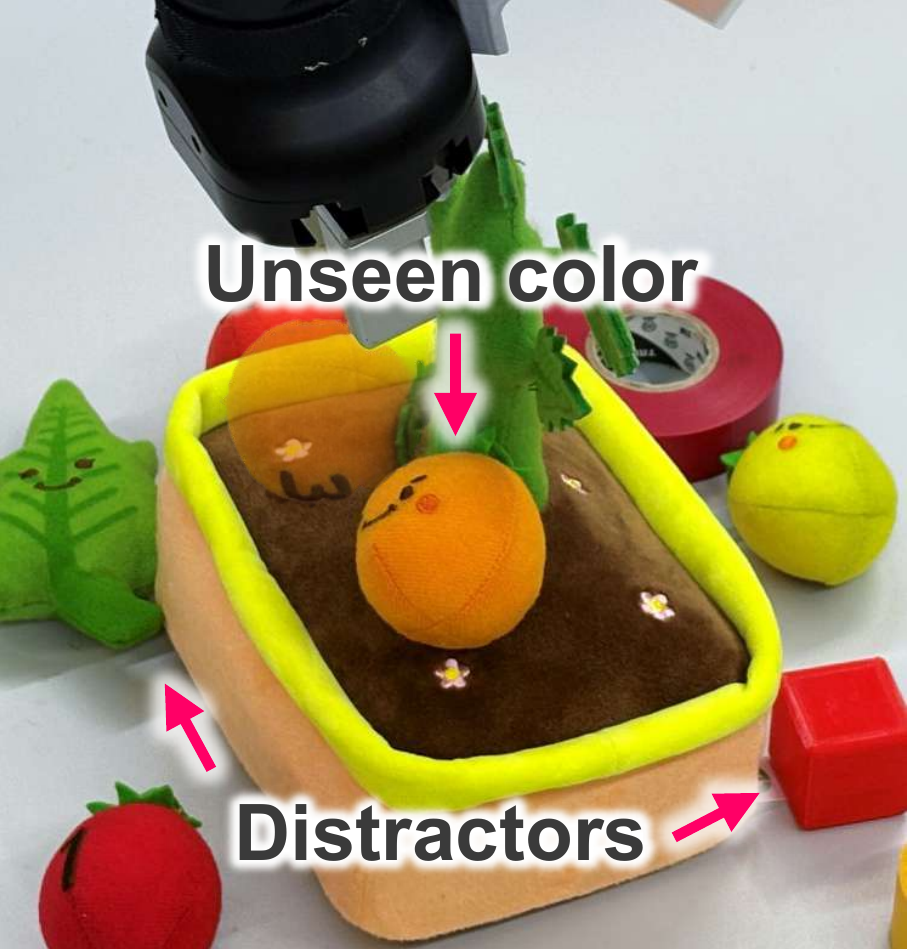}
        \subcaption{Test environment}
        \label{fig:tomato_task_env:test}
    \end{minipage}
    \caption{
    Task environments for the tomato harvesting task.
    (a) shows a demonstration environment. ``G1'' and ``G2'' mean two goals on the task.
    (b) shows a test environment. 
    The test environment has distractors, and the color of the tomatoes changes.
    }
    \label{fig:tomato_task_env}
\end{figure}
\begin{figure}[t]
    \centering
    \begin{minipage}[b]{0.49\linewidth}
        \centering
        \includegraphics[width=0.8\linewidth]{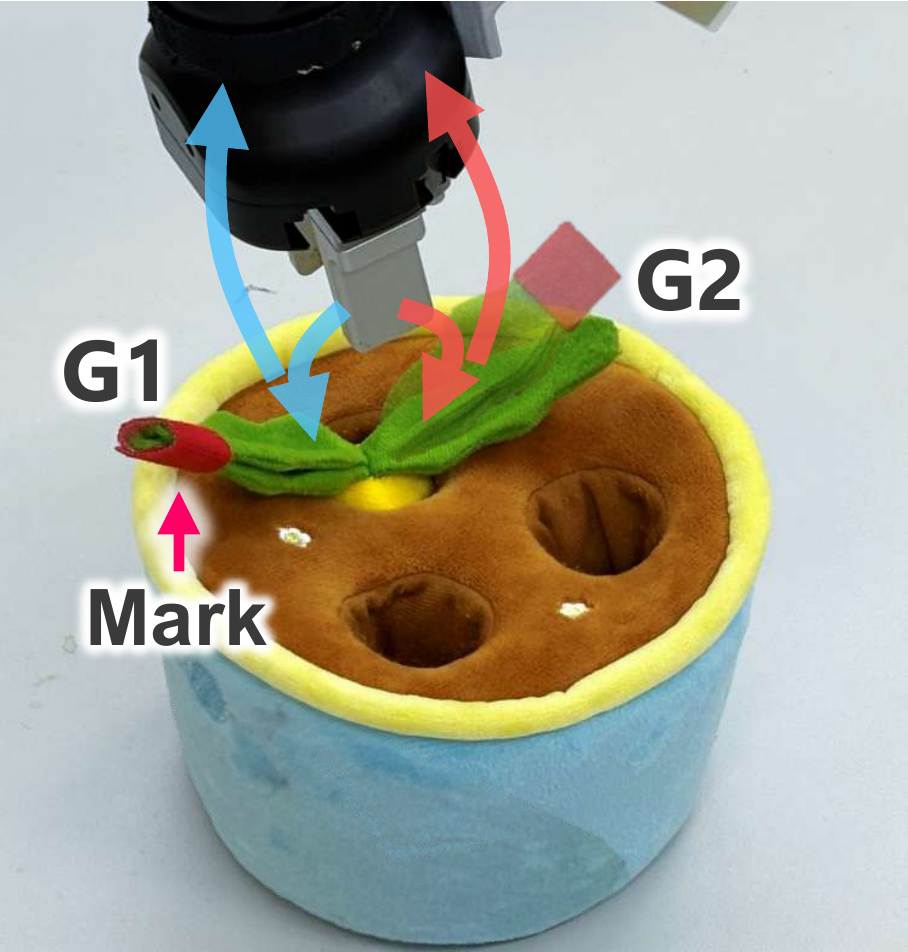}
        \subcaption{Demonstration environment}
        \label{fig:carrot_task_env:demo}
    \end{minipage}
    \begin{minipage}[b]{0.49\linewidth}
        \centering
        \includegraphics[width=0.8\linewidth]{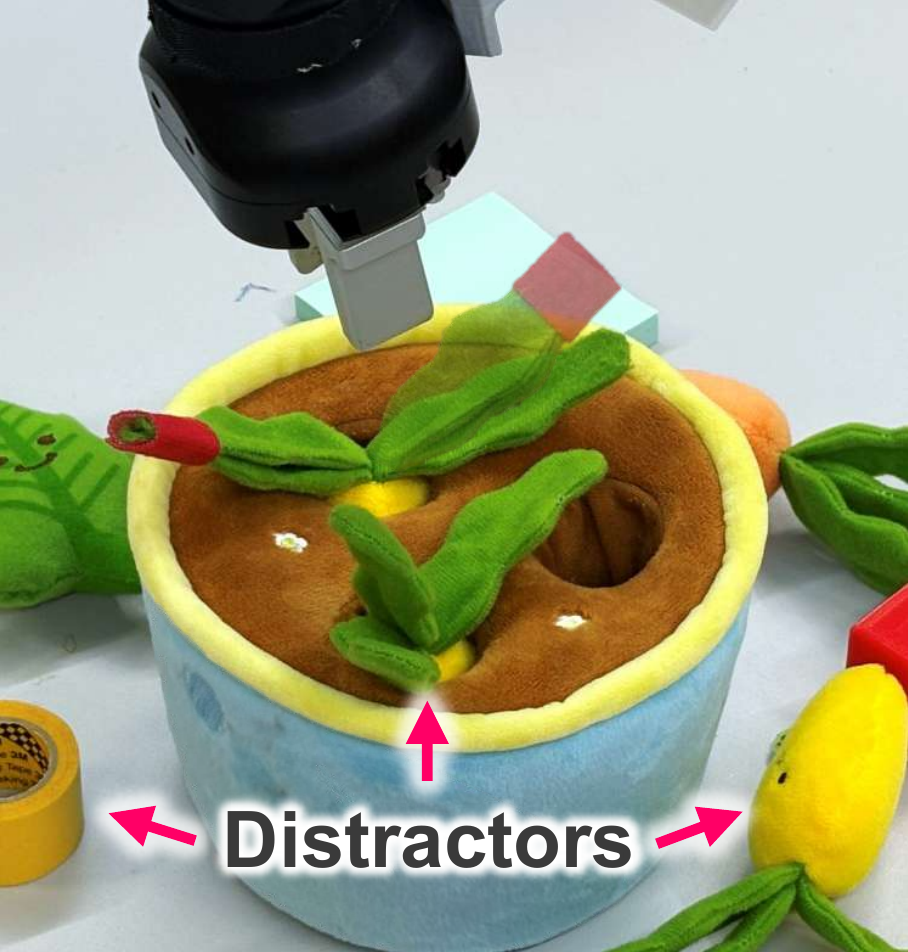}
        \subcaption{Test environment}
        \label{fig:carrot_task_env:test}
    \end{minipage}
    \caption{
    Task environments for the carrot harvesting task
    (a) shows a demonstration environment.
    ``G1'' and ``G2'' mean two goals on the task.
    (b) shows a test environment. 
    The test environment has distractors and another carrot on the pot.
    }
    \label{fig:carrot_task_env}
\end{figure}

\subsection{Representation Generalization Metrics}
In this study, we employ the Absolute RND Gap (ARG) as a quantitative metric to evaluate the visual generalization capability of the encoder in visuomotor policies.
ARG is defined as the difference in Random Network Distillation (RND) values~\cite{burda2019RND} of a trained encoder $E_{\phi^*}$ between the demonstration environment and the test environment.
A smaller ARG indicates higher visual generalization performance.
RND is computed using a pair of neural networks with identical architectures: a fixed network $f_{\theta_\mathrm{fix}}$ and a trainable network $f_{\theta}$.
The fixed network $f_{\theta_\mathrm{fix}}$ has frozen parameters and maps the encoder output to a feature vector.
The trainable network $f_{\theta}$ is optimized to predict the output of the fixed network $f_{\theta_\mathrm{fix}}$, as described below:
\begin{align}
    \theta^* = 
    \mathop{\mathrm{argmin}}_{\theta} 
    \sum_{\mathbf{o} \in \mathcal{D}}
    \left\|
        f_{\theta} (E_{\phi^*}(\mathbf{o})) -
        f_{\theta_{\mathrm{fix}}} (E_{\phi^*}(\mathbf{o})) 
    \right\|_2^2.
\label{eq: ARG_predictor_loss}
\end{align}
RND is calculated using two neural networks as follow:
\begin{align}
    \mathrm{RND}(\mathbf{o}, \theta^*, E_{\phi^*}) = 
    \|
        f_{\theta^*} (E_{\phi^*}(\mathbf{o})) -
        f_{\theta_{\mathrm{fix}}} (E_{\phi^*}(\mathbf{o}))
    \|_2^2.
\end{align}

ARG is computed as the difference between the RND values obtained when evaluating the encoder on visual observations collected in the demonstration environment and those collected in the test environment.
$\mathcal O^\mathrm{d}$ denotes visual observations collected in the demonstration environment, and 
$\mathcal O^\mathrm{t}$ denotes those collected in the test environment.
ARG is then defined as follows:
\begin{align}
    \mathrm{ARG}(E_{\phi^*}) = 
    \left| 
        \mathbb E_{\mathbf{o} \in \mathcal{O}^\mathrm{d}} [\mathrm{RND}(\mathbf{o})] -
        \mathbb E_{\mathbf{o} \in \mathcal{O}^\mathrm{t}} [\mathrm{RND}(\mathbf{o})]
    \right|. 
\end{align}


\begin{figure}[t]
    \centering
    \begin{minipage}[b]{0.49\linewidth}
        \centering
        \includegraphics[width=0.8\linewidth]{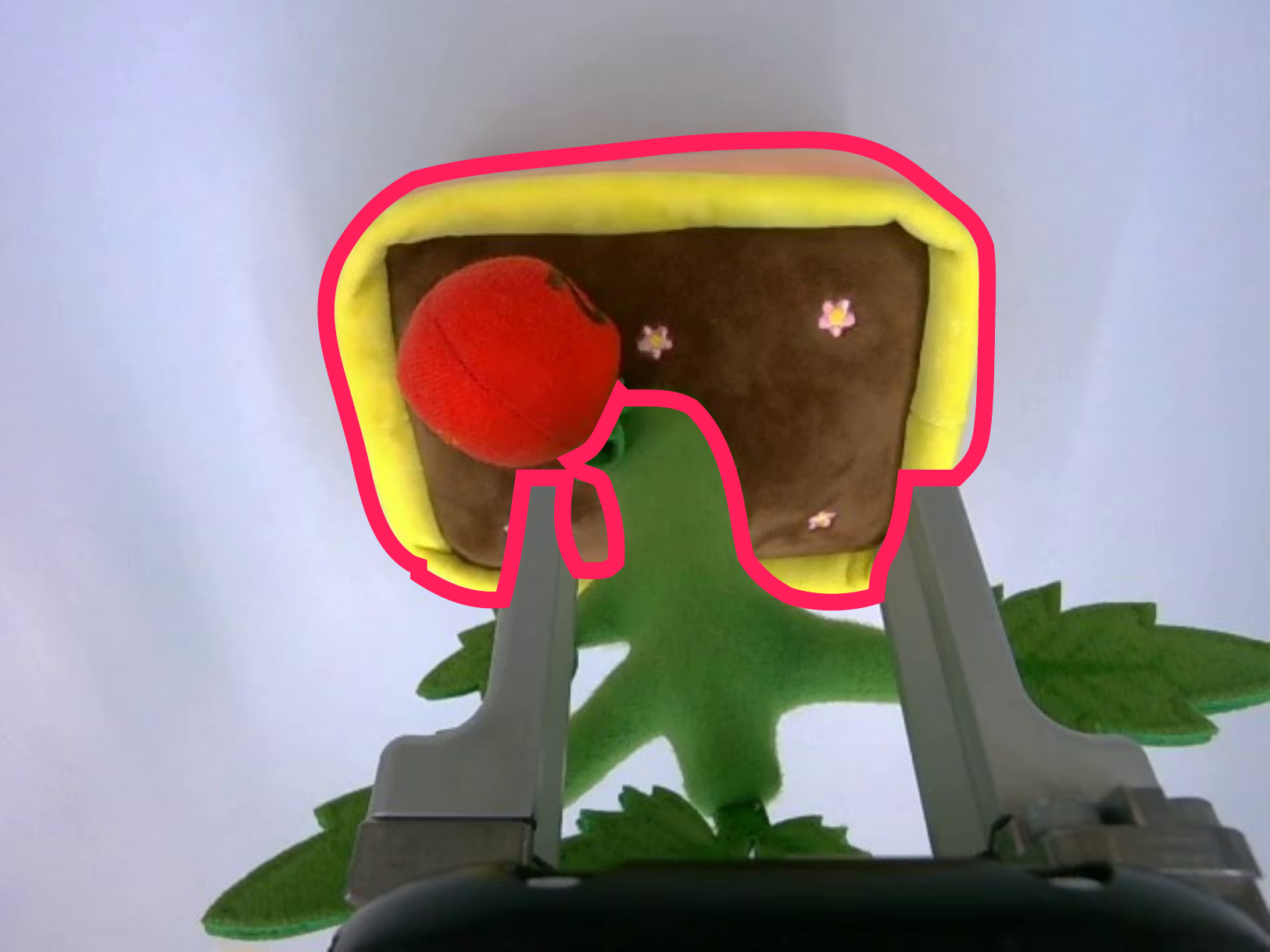}
        \subcaption{Visual observation}
        \label{fig:simple_task:data_augmentation:setting:tomato:visual_observation}
    \end{minipage}
    \begin{minipage}[b]{0.49\linewidth}
        \centering
        \includegraphics[width=0.8\linewidth]{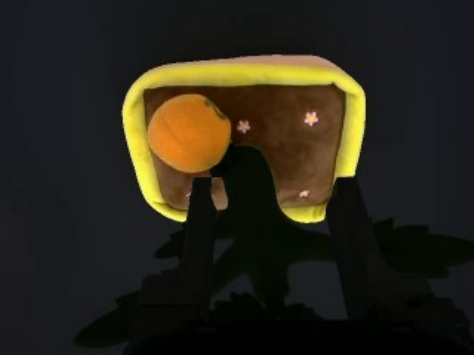}
        \subcaption{Task-relevant aug.}
        \label{fig:simple_task:data_augmentation:setting:tomato:task_relevant_aug}
    \end{minipage}
    \begin{minipage}[b]{0.49\linewidth}
        \centering
        \includegraphics[width=0.8\linewidth]{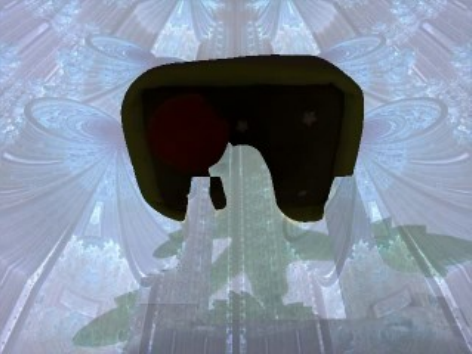}
        \subcaption{Task-irrelevant aug.}
        \label{fig:simple_task:data_augmentation:setting:tomato:task_irrelevant_aug}
    \end{minipage}
    \begin{minipage}[b]{0.49\linewidth}
        \centering
        \includegraphics[width=0.8\linewidth]{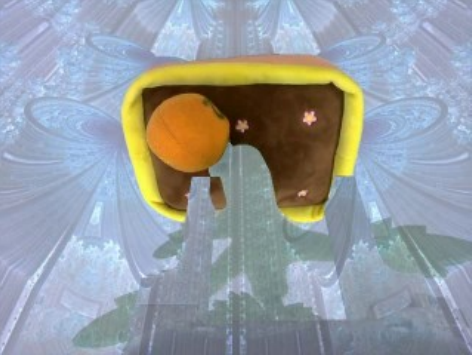}
        \subcaption{Dual augmentation}
        \label{fig:simple_task:data_augmentation:setting:tomato:dual_aug}
    \end{minipage}
    \caption{
    Data augmentation in the tomato harvesting task.
    (a) shows visual observation.
    The red line indicates the borderline of task-relevant and task-irrelevant regions.
    (b) shows task-relevant augmented visual observation.
    (c) shows task-irrelevant augmented visual observation.
    (d) shows dual augmented visual observation.
    }
    \label{fig:simple_task:data_augmentation:setting:tomato}
\end{figure}

\subsection{Experiments with Artificial Vegetable Harvesting Tasks}
\subsubsection{Settings}
We apply DRAIL to two harvesting tasks using artificial vegetables to validate its effectiveness.
\figref{fig:tomato_task_env} and \figref{fig:carrot_task_env} show overview of the tomato and carrot harvesting tasks, respectively.
The tomato harvesting task aims to pinch and lift a tomato, and the carrot harvesting task aims to pinch a marked carrot leaf and pull the carrot out.
These tasks have two goals, and the robot must move from the camera mounted on the gripper toward an appropriate grasp target.
We define task success as whether the robot successfully grasps the target object after executing the motion.
As test environments, for the tomato harvesting task, we replace the red tomato with an orange or yellow one, as shown in \figref{fig:tomato_task_env:test}, and for the carrot harvesting task, we place an additional carrot in the pot besides the grasp target, as shown in \figref{fig:carrot_task_env:test}.
In each test environment, the background contains objects that are irrelevant to the task.

\begin{figure}[t]
    \centering
    \begin{minipage}[b]{0.49\linewidth}
        \centering
        \includegraphics[width=0.8\linewidth]{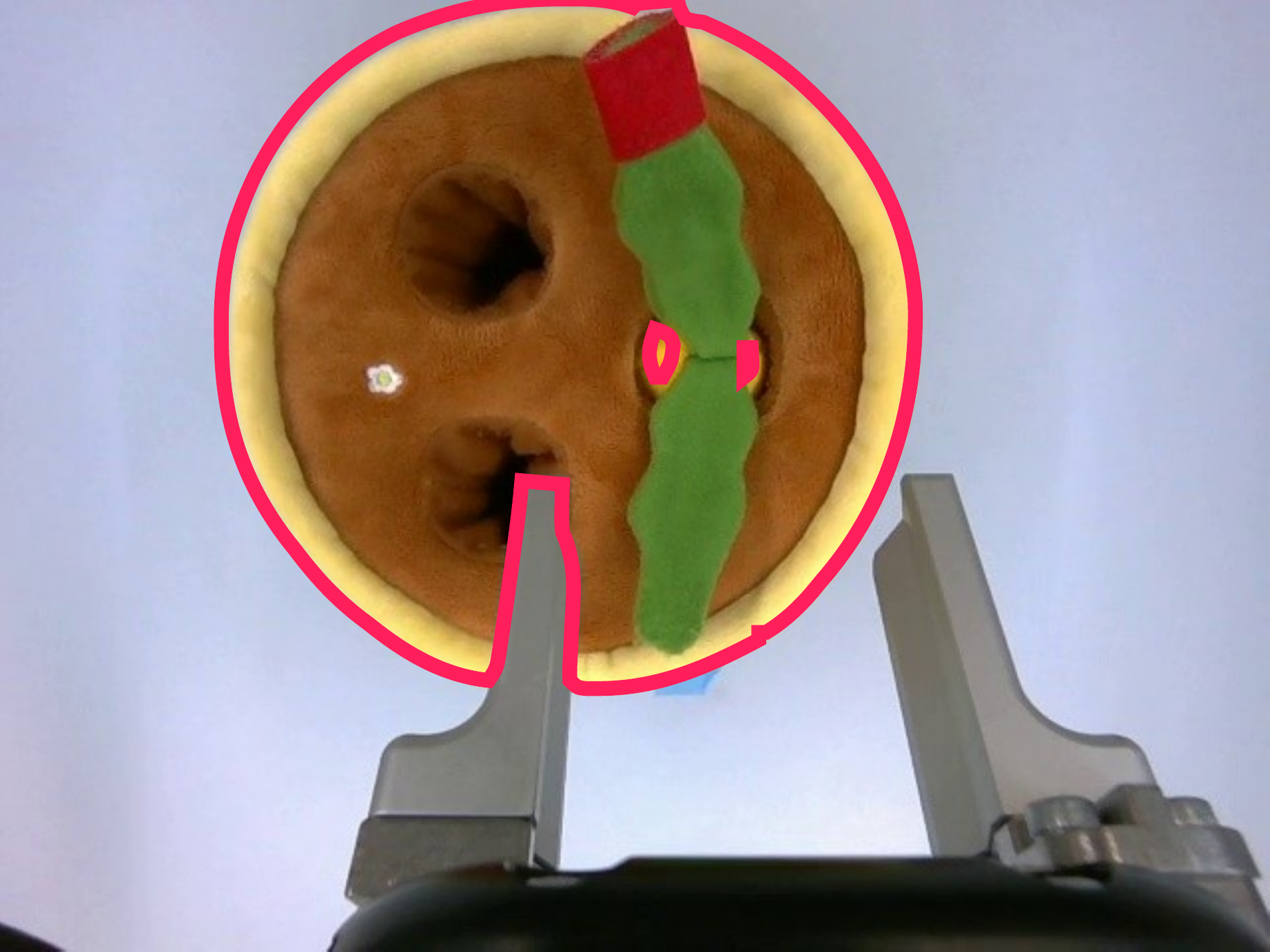}
        \subcaption{Visual observation}
        \label{fig:simple_task:data_augmentation:setting:carrot:visual_observation}
    \end{minipage}
    \begin{minipage}[b]{0.49\linewidth}
        \centering
        \includegraphics[width=0.8\linewidth]{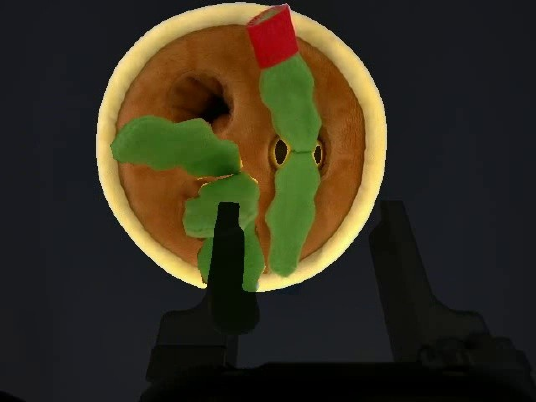}
        \subcaption{Task-relevant aug.}
        \label{fig:simple_task:data_augmentation:setting:carrot:task_relevant_aug}
    \end{minipage}
    \begin{minipage}[b]{0.49\linewidth}
        \centering
        \includegraphics[width=0.8\linewidth]{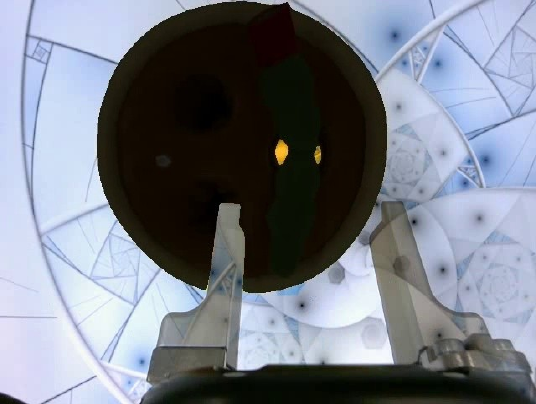}
        \subcaption{Task-irrelevant aug.}
        \label{fig:simple_task:data_augmentation:setting:carrot:task_irrelevant_aug}
    \end{minipage}
    \begin{minipage}[b]{0.49\linewidth}
        \centering
        \includegraphics[width=0.8\linewidth]{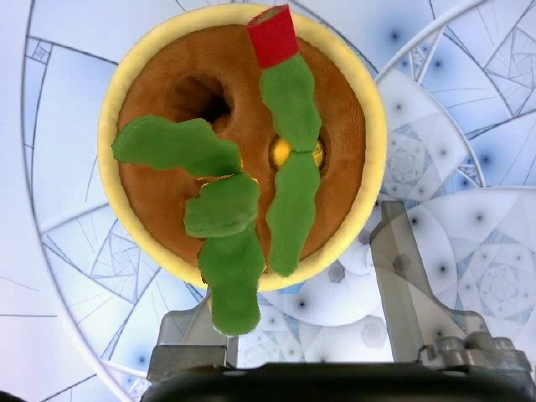}
        \subcaption{Dual augmentation}
        \label{fig:simple_task:data_augmentation:setting:carrot:dual_aug}
    \end{minipage}
    \caption{
    Data augmentation in the carrot harvesting task.
    (a) shows visual observation.
    The red line indicates the borderline of task-relevant and task-irrelevant regions.
    (b) shows task-relevant augmented visual observation.
    (c) shows task-irrelevant augmented visual observation.
    (d) shows dual augmented visual observation.
    }
    \label{fig:simple_task:data_augmentation:setting:carrot}
\end{figure}

In both tasks, we define the task-relevant region as the tomato/carrot and the interior of the pot.
As task-specific augmentation, for the tomato harvesting task, we change the tomato color to yellow-green or orange, and for the carrot harvesting task, we composite cutout leaf images into the interior of the pot at various angles and scales.
Examples of augmented images for the two tasks are shown in \figref{fig:simple_task:data_augmentation:setting:tomato} and \figref{fig:simple_task:data_augmentation:setting:carrot}.
We collect 40 demonstration episodes in the demonstration environment to train policies.
For ARG evaluation, we also collect 20 demonstration episodes in the test environment.
\begin{figure}[!t]
    \centering
    \begin{minipage}[b]{0.9\linewidth}
        \centering
        \includegraphics[width=\linewidth]{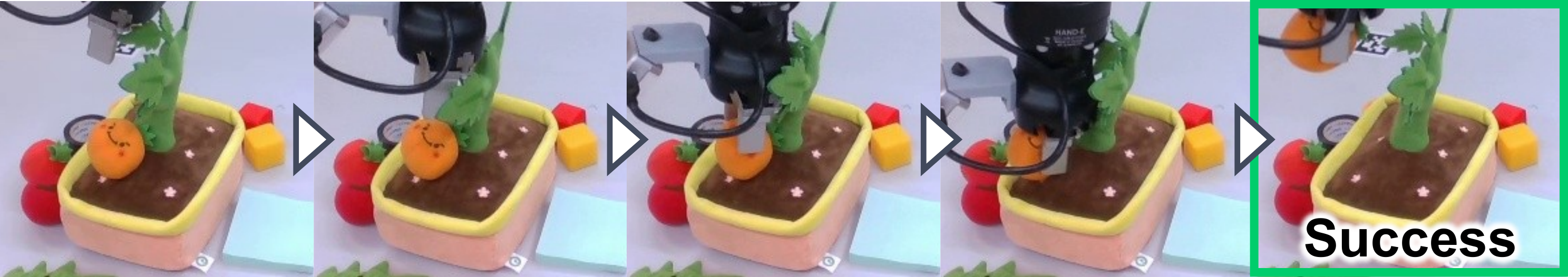}
        \subcaption{DRAIL}
        \label{fig:tomato_task_robot_motion_drail}
    \end{minipage}
    \begin{minipage}[b]{0.9\linewidth}
        \centering
        \includegraphics[width=\linewidth]{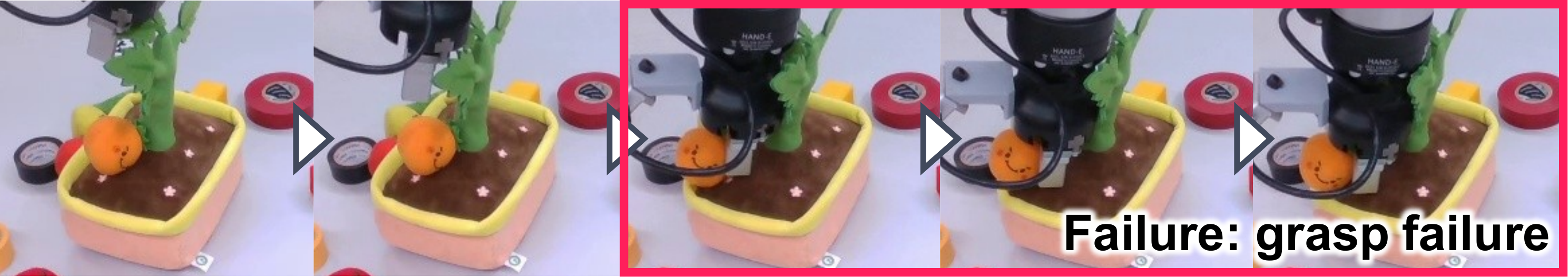}
        \subcaption{DRAIL w/o task-irr. aug.}
        \label{fig:tomato_task_robot_motion_rel}
    \end{minipage}
    \begin{minipage}[b]{0.9\linewidth}
        \centering
        \includegraphics[width=\linewidth]{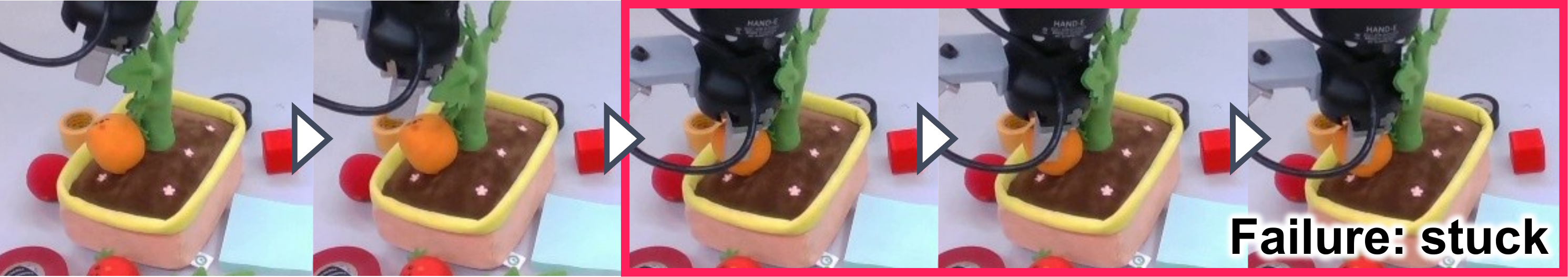}
        \subcaption{DRAIL w/o task-rel. aug.}
        \label{fig:tomato_task_robot_motion_ishida}
    \end{minipage}
    \begin{minipage}[b]{0.9\linewidth}
        \centering
        \includegraphics[width=\linewidth]{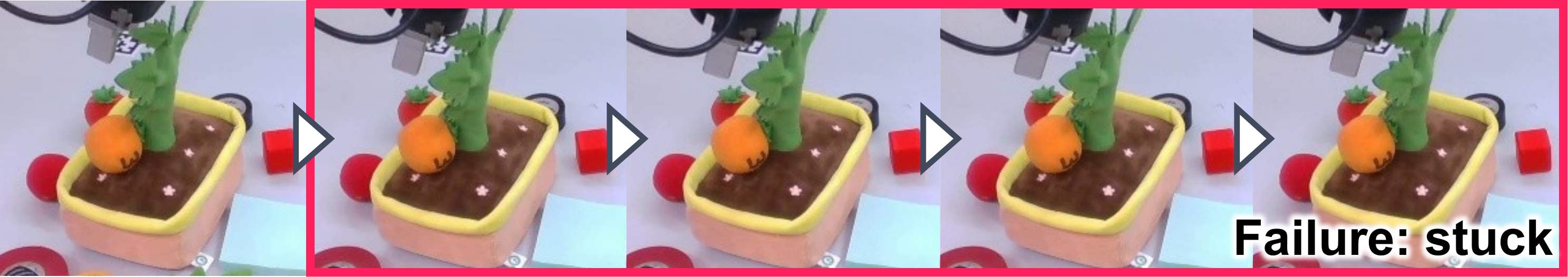}
        \subcaption{DRAIL w/o dual aug.}
        \label{fig:tomato_task_robot_motion_dp}
    \end{minipage}
    \caption{Robot motion using policy learned by DRAIL and ablation methods in the tomato harvesting task}
    \label{fig:tomato_task_robot_motion}
\end{figure}
\begin{figure}[t]
    \centering
    \begin{minipage}[b]{0.9\linewidth}
        \centering
        \includegraphics[width=\linewidth]{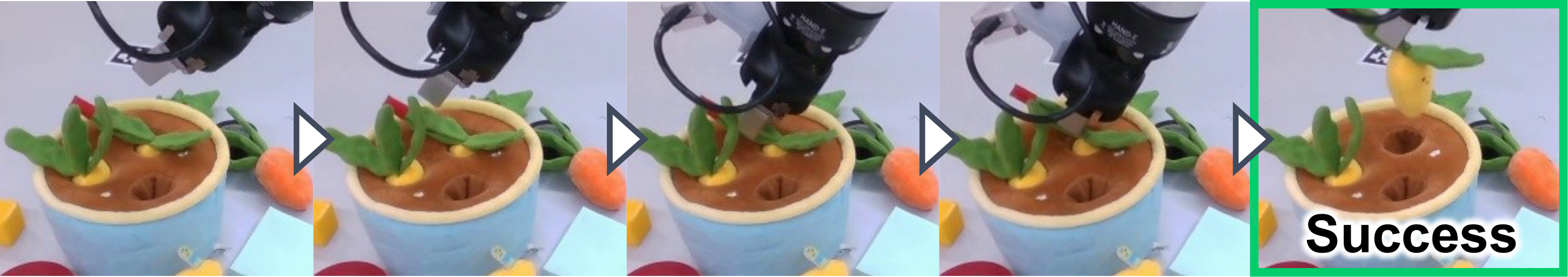}
        \subcaption{DRAIL}
        \label{fig:carrot_task_robot_motion_drail}
    \end{minipage}
    \begin{minipage}[b]{0.9\linewidth}
        \centering
        \includegraphics[width=\linewidth]{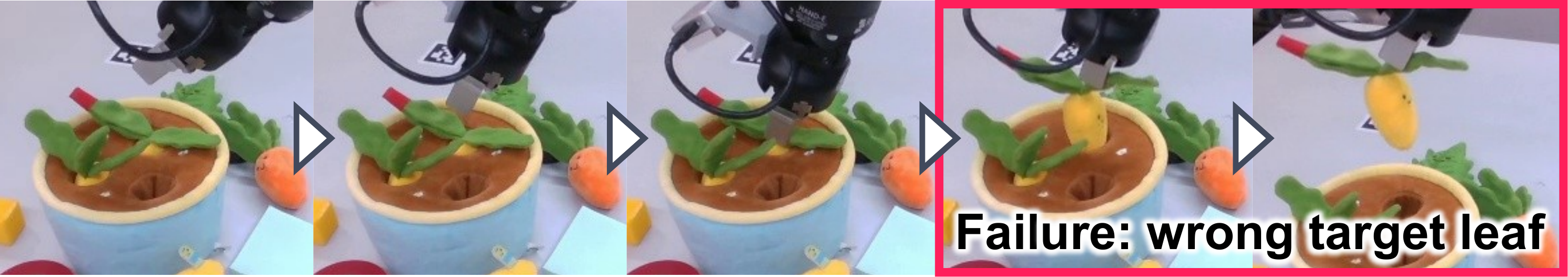}
        \subcaption{DRAIL w/o task-irr. aug.}
        \label{fig:carrot_task_robot_motion_rel}
    \end{minipage}
    \begin{minipage}[b]{0.9\linewidth}
        \centering
        \includegraphics[width=\linewidth]{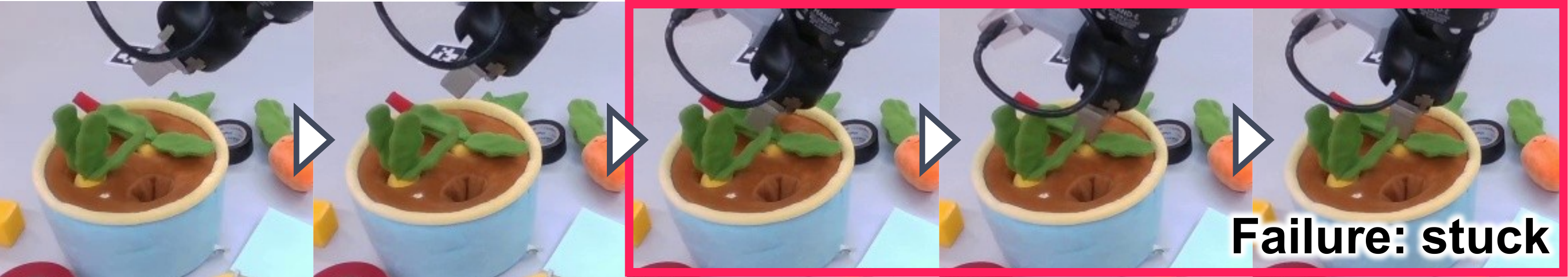}
        \subcaption{DRAIL w/o task-rel. aug.}
        \label{fig:carrot_task_robot_motion_ishida}
    \end{minipage}
    \begin{minipage}[b]{0.9\linewidth}
        \centering
        \includegraphics[width=\linewidth]{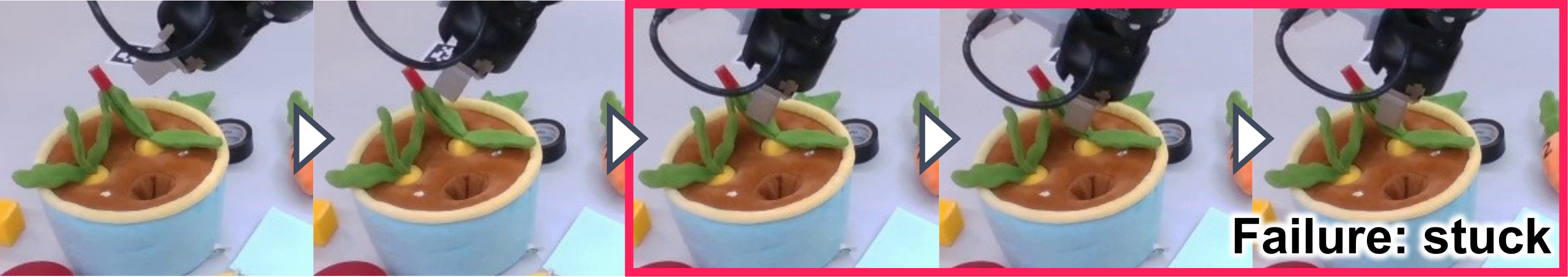}
        \subcaption{DRAIL w/o dual aug.}
        \label{fig:carrot_task_robot_motion_dp}
    \end{minipage}%
    \caption{Robot motion using policy learned by DRAIL and ablation methods in the carrot harvesting task}
    \label{fig:carrot_task_robot_motion}
\end{figure}

\begin{table}[t]
    \centering
    \caption{
    Success rates in the tomato harvesting task.
    10 trials were performed for each condition.
    }
    \label{tab:toy_task:success_rate:tomato}
    \begin{tabular}{cc|cccc}
        \toprule
        Env. & Goal & DRAIL          & DRAIL           & DRAIL          & DRAIL \\
             &      &                & w/o task-irr.   & w/o task-rel.  & w/o dual  \\
        \midrule                                                    
        \midrule                                    
        Demo & G1   & \textbf{100\%} & \textbf{100\%}  & \textbf{100\%} & \textbf{100\%} \\
        Demo & G2   & \textbf{100\%} & \textbf{100\%}  & \textbf{100\%} & \textbf{100\%} \\
        \midrule                                                    
        Test & G1   & \textbf{100\%} & 70\%            & 0\%            & 0\%            \\
        Test & G2   & \textbf{100\%} & 90\%            & 0\%            & 0\%            \\
        \bottomrule
    \end{tabular}
    \vspace{5mm}
    \centering
    \caption{
    Success rates in the carrot vegetable harvesting task.
    10 trials were performed for each condition.
    }
    \label{tab:toy_task:success_rate:carrot}
    \begin{tabular}{cc|cccc}
        \toprule
        Env. & Goal & DRAIL           & DRAIL          & DRAIL           & DRAIL \\
             &      &                 & w/o task-irr.  & w/o task-rel.   & w/o dual  \\
        \midrule                                    
        \midrule                                    
        Demo & G1   & \textbf{100\%}  & \textbf{100\%} & 90\%            & \textbf{100\%} \\
        Demo & G2   & \textbf{100\%}  & \textbf{100\%} & \textbf{100\%}  & 80\%           \\
        \midrule                                                    
        Test & G1   & \textbf{100\%}  & 0\%            & 0\%             & 0\%            \\
        Test & G2   & \textbf{90\%}   & 70\%           & 40\%            & 0\%            \\
        \bottomrule
    \end{tabular}
\end{table}
\begin{figure}[!t]
    \centering
    \includegraphics[width=0.9\linewidth]{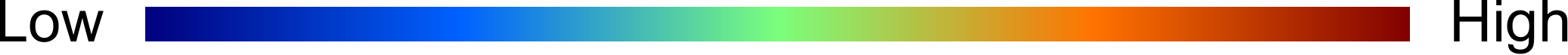}
    \begin{minipage}[b]{0.45\linewidth}
        \centering
        \includegraphics[width=0.8\linewidth]{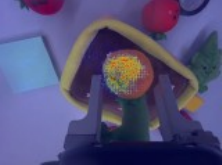}
        \subcaption{DRAIL}
        \label{fig:task_activation_map:tomato:drail}
    \end{minipage}
    \begin{minipage}[b]{0.45\linewidth}
        \centering
        \includegraphics[width=0.8\linewidth]{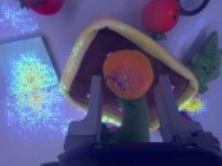}
        \subcaption{DRAIL w/o task-irr. aug.}
        \label{fig:task_activation_map:tomato:rel}
    \end{minipage} 
    \begin{minipage}[b]{0.45\linewidth}
        \centering
        \includegraphics[width=0.8\linewidth]{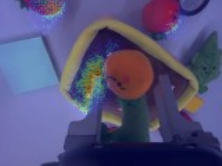}
        \subcaption{DRAIL w/o task-rel. aug.}
        \label{fig:task_activation_map:tomato:ishida}
    \end{minipage} 
    \begin{minipage}[b]{0.45\linewidth}
        \centering
        \includegraphics[width=0.8\linewidth]{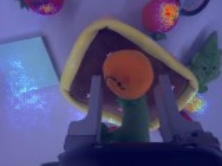}
        \subcaption{DRAIL w/o dual aug.}
        \label{fig:task_activation_map:tomato:dp}
    \end{minipage}
    \begin{minipage}[b]{0.45\linewidth}
        \centering
        \includegraphics[width=0.8\linewidth]{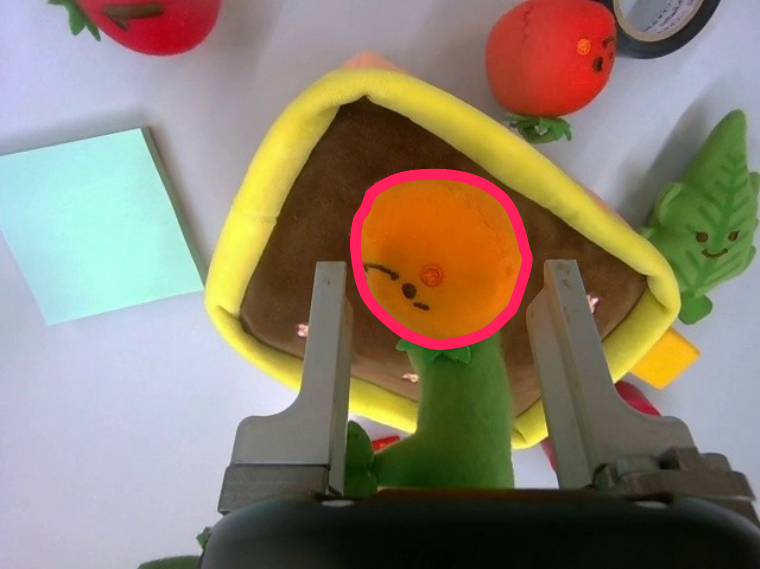}
        \subcaption{Visual observation}
        \label{fig:tomato_task_activation_map_original}
    \end{minipage}
    \caption{
    Visualization of the attention region of the learned policy using the saliency map in the tomato harvesting task. 
    In the visual observation shown in (e), the red line highlights the important region that the policy should attend to for the task.
    }
    \label{fig:task_activation_map:tomato}
\end{figure}

\subsubsection{Results}
To answer RQ1, we evaluate policies based on task success rates in the demonstration and test environments.
Policy motions in the tomato harvesting task and the carrot harvesting task are shown in \figref{fig:tomato_task_robot_motion} and \figref{fig:carrot_task_robot_motion}, respectively.
The corresponding success rates are reported in \tabref{tab:toy_task:success_rate:tomato} and \tabref{tab:toy_task:success_rate:carrot}, respectively.
In the demonstration environment, all methods achieve consistently high success rates across both goal settings.
In contrast, the ablation methods exhibit a substantial performance degradation in the test environment, with particularly severe degradation for methods without task-relevant augmentation.
These results demonstrate that robustness to visual diversity is only achieved using both augmentations of DRAIL.

\begin{figure}[t]
    \centering
    \includegraphics[width=0.9\linewidth]{figures/colorbar.pdf}
    \begin{minipage}[b]{0.45\linewidth}
        \centering
        \includegraphics[width=0.8\linewidth]{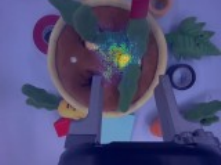}
        \subcaption{DRAIL}
        \label{fig:carrot_task_activation_map_drail}
    \end{minipage}
    \begin{minipage}[b]{0.45\linewidth}
        \centering
        \includegraphics[width=0.8\linewidth]{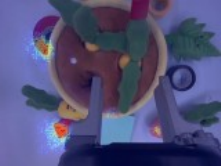}
        \subcaption{DRAIL w/o task-irr. aug.}
        \label{fig:carrot_task_activation_map_rel}
    \end{minipage} 
    \begin{minipage}[b]{0.45\linewidth}
        \centering
        \includegraphics[width=0.8\linewidth]{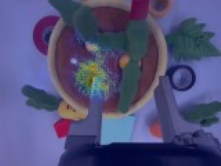}
        \subcaption{DRAIL w/o task-rel. aug.}
        \label{fig:carrot_task_activation_map_ishida}
    \end{minipage} 
    \begin{minipage}[b]{0.45\linewidth}
        \centering
        \includegraphics[width=0.8\linewidth]{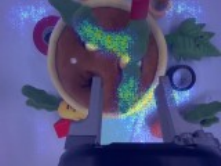}
        \subcaption{DRAIL w/o dual aug.}
        \label{fig:carrot_task_activation_map_dp}
    \end{minipage}
    \begin{minipage}[b]{0.45\linewidth}
        \centering
        \includegraphics[width=0.8\linewidth]{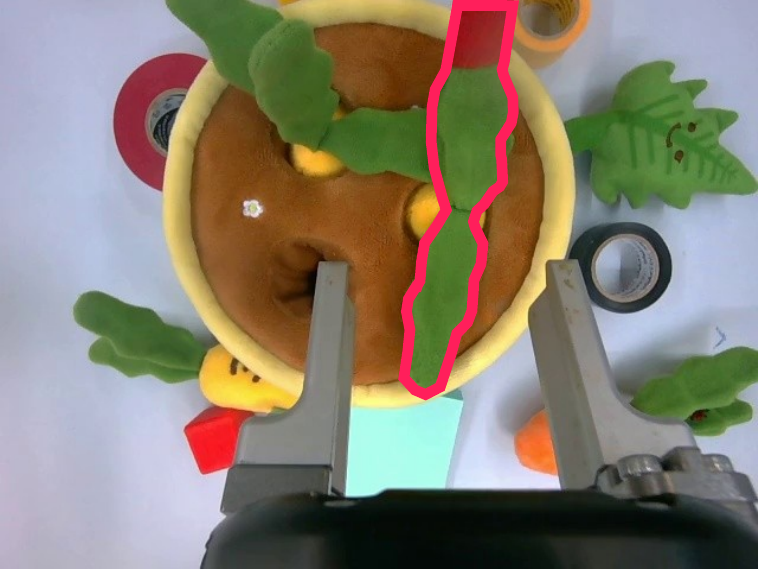}
        \subcaption{Visual observation}
        \label{fig:carrot_task_activation_map_original}
    \end{minipage} 
    \caption{
    Visualization of the attention region of the learned policy by the saliency map in the carrot harvesting task.
    In the visual observation shown in (e), the red line highlights the important region that the policy should attend to for the task.
    }
    \label{fig:task_activation_map:carrot}
\end{figure}

To address RQ2, we visualize attention regions of policies in the test environment.
Attention regions are shown in \figref{fig:task_activation_map:tomato} for the tomato harvesting task and in \figref{fig:task_activation_map:carrot} for the carrot harvesting task.
In both tasks, DRAIL consistently focuses attention on the target crop, whereas the ablation methods cannot.
For DRAIL without task-irrelevant augmentation, the policy's attention spreads in background regions.
Conversely, for DRAIL without task-relevant augmentation, the policy's attention is directed toward non-crop regions within the pot, rather than the crop itself.
These results indicate that our dual data augmentation effectively guides the policy's attention toward the regions that should be attended to for the task.

\begin{table}[t]
    \centering
    \caption{
    Comparison of ARGs of policies learned by each method in the tomato harvesting task.
    For each method, ARG was recomputed 20 times with different random samplings of the fixed network.
    Statistical significance was assessed using a paired t-test against DRAIL. 
    ``$\dagger$" indicates a statistically significant difference with $p < 0.05$.
    }
    \label{tab:toy_task:ARG_result:tomato}
    \begin{tabular}{c|cccc}
        \toprule
                    & DRAIL                      & DRAIL              & DRAIL             & DRAIL \\
        ARG         &                            & w/o task-irr.      & w/o task-rel.     & w/o dual  \\
        \midrule                                                      
        \midrule                                                      
        Mean      & $\mathbf{7.8\times 10^3}$ & $2.5\times 10^{4\dagger}$  & $2.4 \times 10^{6\dagger}$ & $1.2\times 10^{6\dagger}$  \\
        STD       & $\mathbf{6.0\times 10^3}$ & $2.1\times 10^4$~  & $4.9 \times 10^5$~ & $3.0\times 10^5$~  \\
        \bottomrule
    \end{tabular}
\end{table}
\begin{table}[t]
    \centering
    \caption{
    Comparison of ARGs of policies learned by each method in the carrot harvesting task.
    For each method, ARG was recomputed 20 times with different random samplings of the fixed network.
    Statistical significance was assessed using a paired t-test against DRAIL. 
    ``$\dagger$" indicates a statistically significant difference with $p < 0.05$.
    }
    \label{tab:toy_task:ARG_result:carrot}
    \begin{tabular}{c|cccc}
        \toprule
                    & DRAIL                      & DRAIL              & DRAIL             & DRAIL \\
        ARG         &                            & w/o task-irr.      & w/o task-rel.     & w/o dual  \\
        \midrule                                                      
        \midrule                                                      
        Mean        & $\mathbf{1.0 \times 10^4}$ & $7.2 \times 10^{4\dagger}$  & $1.1 \times 10^{5\dagger}$ & $4.5 \times 10^{5\dagger}$  \\
        STD         & $\mathbf{7.3 \times 10^3}$ & $2.3 \times 10^4$~  & $3.3 \times 10^4$~ & $8.6 \times 10^4$~  \\
        \bottomrule
    \end{tabular}
\end{table}

To answer RQ3, the visual generalization of the learned policies is quantified using ARG, and the results are shown in \tabref{tab:toy_task:ARG_result:tomato} and \tabref{tab:toy_task:ARG_result:carrot}.
DRAIL has the lowest ARG of the ablation methods in both tasks.
This suggests that DRAIL's data augmentation allows the image encoder to extract consistent features from visual observations in the test environment and the demo data.
Overall, these results indicate that DRAIL improves generalization in tasks involving appearance diversity in color and shape.

Based on the above results, RQ1-RQ3 were all validated for artificial vegetable harvesting tasks.
DRAIL maintains high success rates even in the test environment.
This is because its data augmentation guides the policy's attention toward semantically meaningful regions and enables the extraction of features that remain consistent under visual variations.

\subsection{Experiment with Real Lettuce Defective Leaf Picking Preparation Task}
\subsubsection{Settings}

\begin{figure}[!t] 
    \centering
    \includegraphics[width=0.6\linewidth]{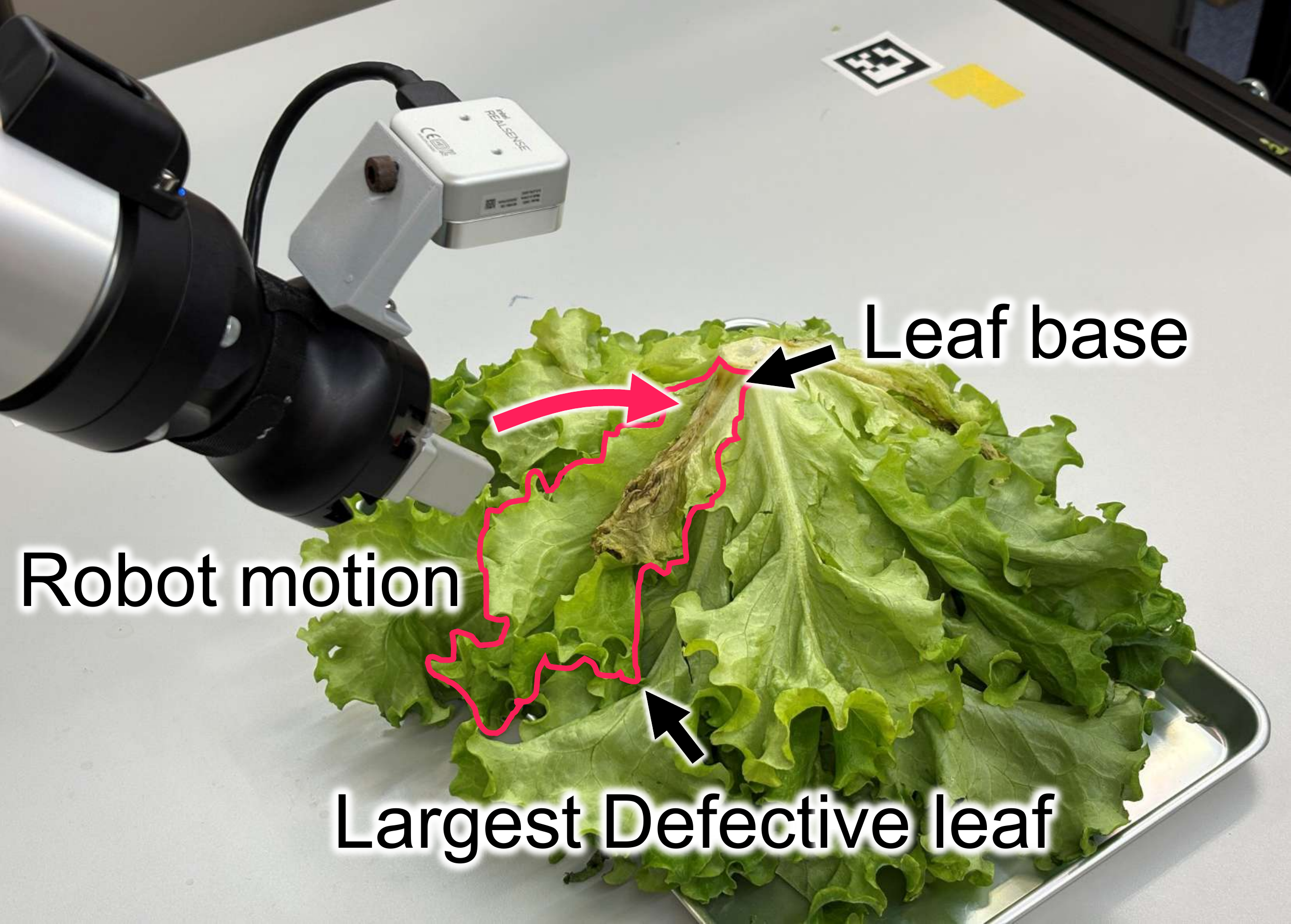}
    \caption{
    Overview of the real lettuce defective leaf picking preparation task.
    Given a top-down wrist-camera observation, the policy drives the gripper toward the leaf base of the most defective leaf among the front-side leaves.
    The target leaf is outlined in red, and the desired grasp point is its leaf base.
    }
    \label{fig:real_lettuce_task_env}
\end{figure}
\begin{figure}[t] 
    \centering
    \begin{minipage}[b]{0.49\linewidth}
        \centering
        \includegraphics[width=0.8\linewidth]{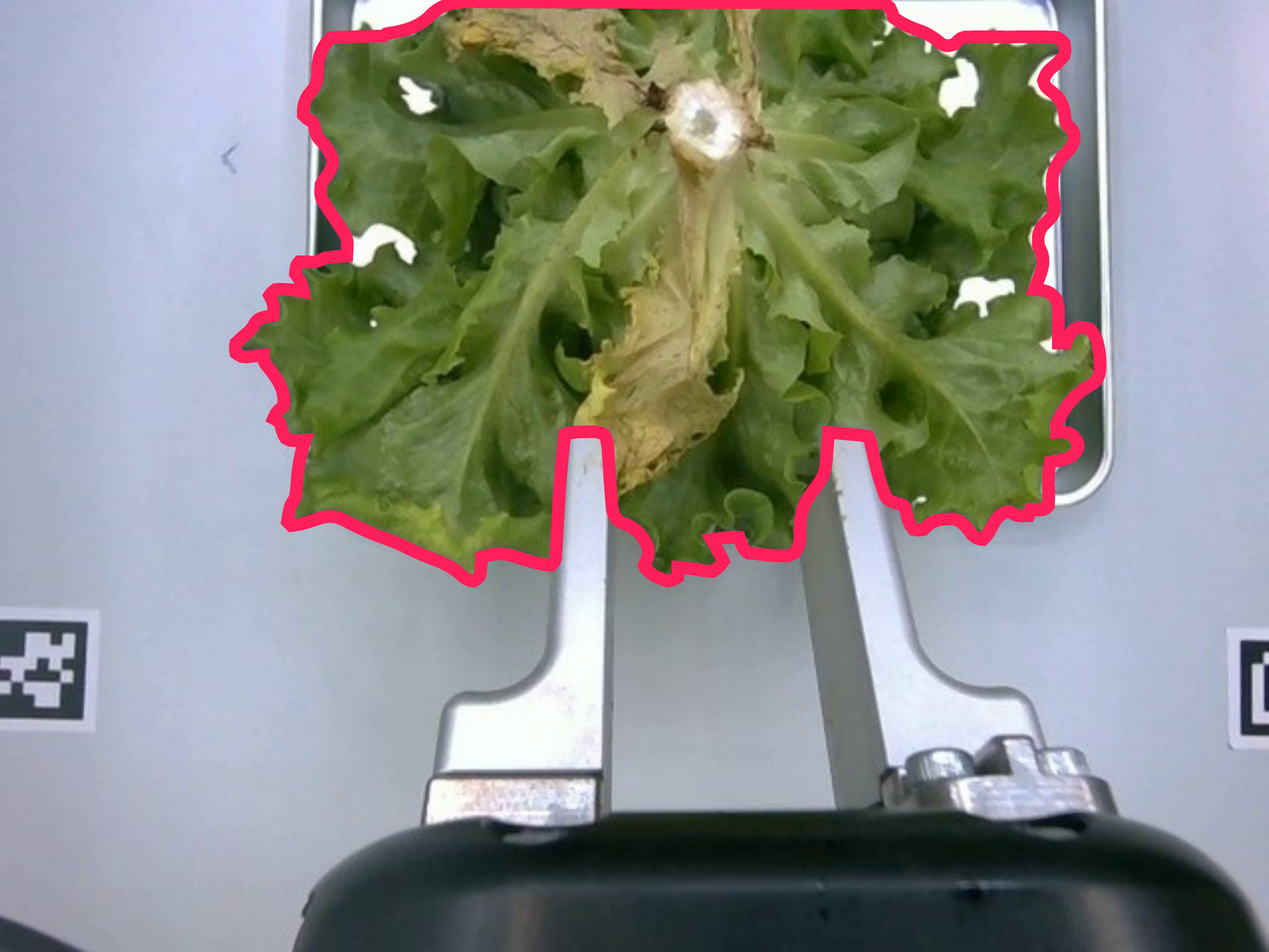}
        \subcaption{Visual observation}
    \end{minipage}
    \begin{minipage}[b]{0.49\linewidth}
        \centering
        \includegraphics[width=0.8\linewidth]{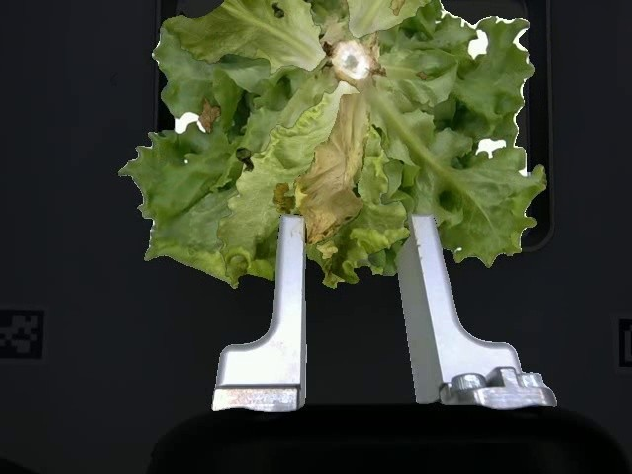}
        \subcaption{Task-relevant aug.}
    \end{minipage}
    \begin{minipage}[b]{0.49\linewidth}
        \centering
        \includegraphics[width=0.8\linewidth]{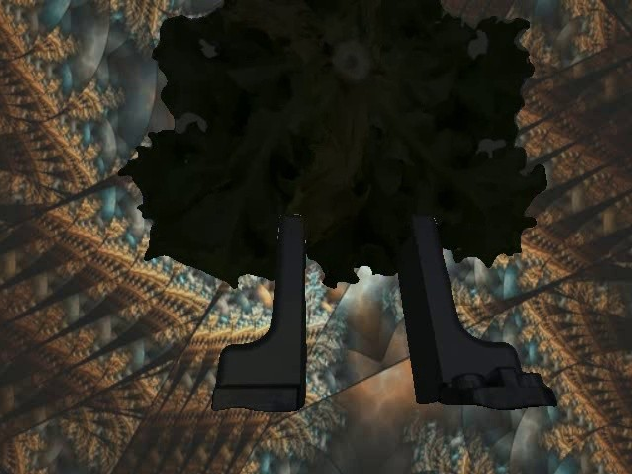}
        \subcaption{Task-irrelevant aug.}
    \end{minipage}
    \begin{minipage}[b]{0.49\linewidth}
        \centering
        \includegraphics[width=0.8\linewidth]{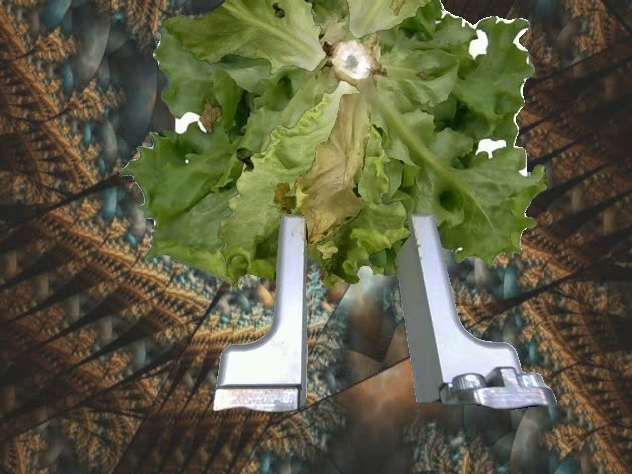}
        \subcaption{Dual augmentation}
        \label{fig:real_lettuce_aug_drail}
    \end{minipage}
    \caption{
    Data augmentation in the real lettuce defective leaf picking preparation task.
    (a) shows visual observation.
    The red line indicates the borderline of task-relevant and task-irrelevant regions.
    (b) shows task-relevant augmented visual observation.
    (c) shows task-irrelevant augmented visual observation.
    (d) shows dual augmented visual observation.
    }
    \label{fig:real_lettuce_aug}
\end{figure}

We evaluate DRAIL on the real lettuce defective leaf picking preparation task, a sequential task performed prior to defective leaf removal.
It consists of two consecutive subtasks: (i) reaching, in which the manipulator approaches the detected defective leaf, and (ii) a supplementary subtask that prepares for stable grasping through fine positioning or alignment.
The task overview is shown in \figref{fig:real_lettuce_task_env}.

Success is measured using two criteria: leaf selection and positional alignment.
Leaf selection indicates whether the policy approaches the appropriate leaf, i.e., whether the leaf closest to the gripper is the one with the largest defect.
Positional alignment indicates whether the gripper opening contains the base of the leaf with the largest defect.

For this task, we define the task-relevant region as the lettuce and gripper, and the remaining area as the task-irrelevant region.
Details of the task-relevant and irrelevant augmentation are shown in \figref{fig:real_lettuce_aug}.
Task-relevant augmentation is implemented by compositing lettuce leaves with no or small defects into the lettuce on the visual observation.
Different lettuces are used in the demonstration and test environments to evaluate robustness to appearance variation.
We collect 23 demonstration episodes in the demonstration environment to train policies.
For ARG evaluation, we also collect 70 demonstration episodes in the test environment.

\subsubsection{Results}
\begin{figure}[t] 
    \centering
    \begin{minipage}[b]{\linewidth}
        \centering
        \includegraphics[width=0.9\linewidth]{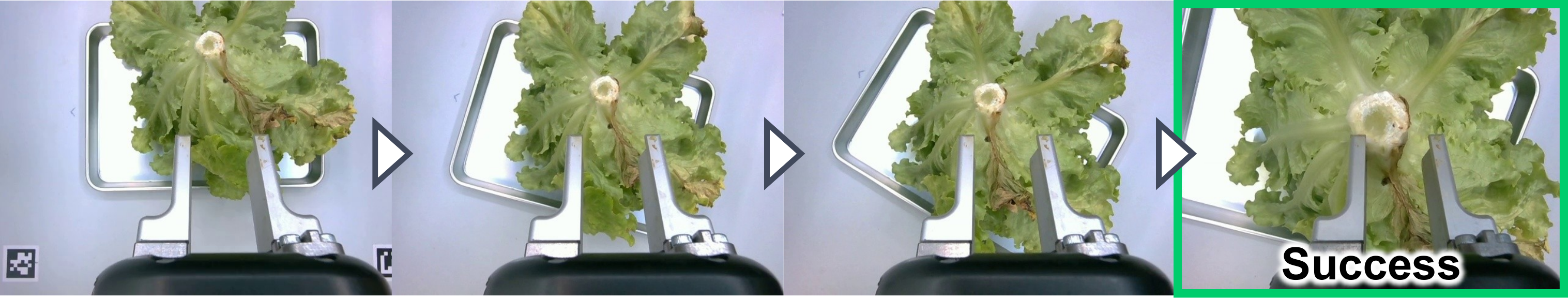}
        \subcaption{DRAIL}
        \label{fig:real_lettuce_task_robot_motion_drail}
    \end{minipage}
    \begin{minipage}[b]{\linewidth}
        \centering
        \includegraphics[width=0.9\linewidth]{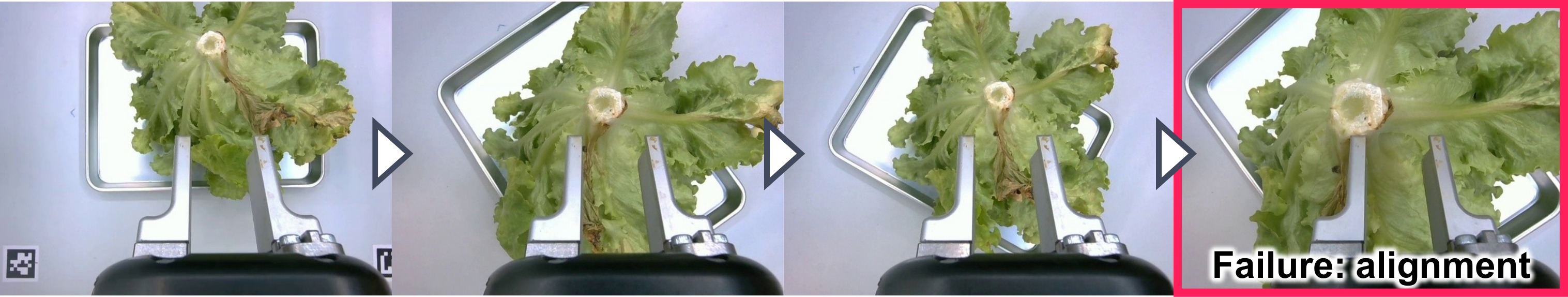}
        \subcaption{DRAIL w/o task-irr. aug.}
        \label{fig:real_lettuce_task_robot_motion_rel}
    \end{minipage}
    \begin{minipage}[b]{\linewidth}
        \centering
        \includegraphics[width=0.9\linewidth]{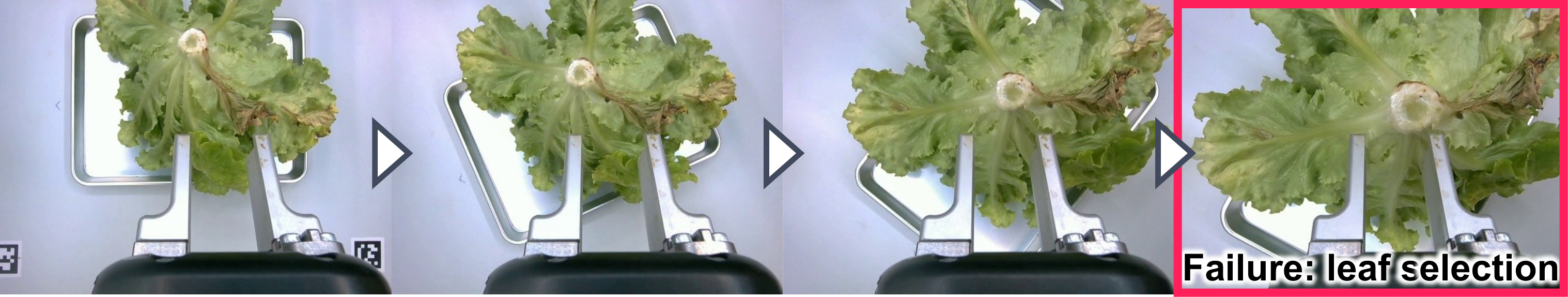}
        \subcaption{DRAIL w/o task-rel. aug.}
        \label{fig:real_lettuce_task_robot_motion_ishida}
    \end{minipage}
    \begin{minipage}[b]{\linewidth}
        \centering
        \includegraphics[width=0.9\linewidth]{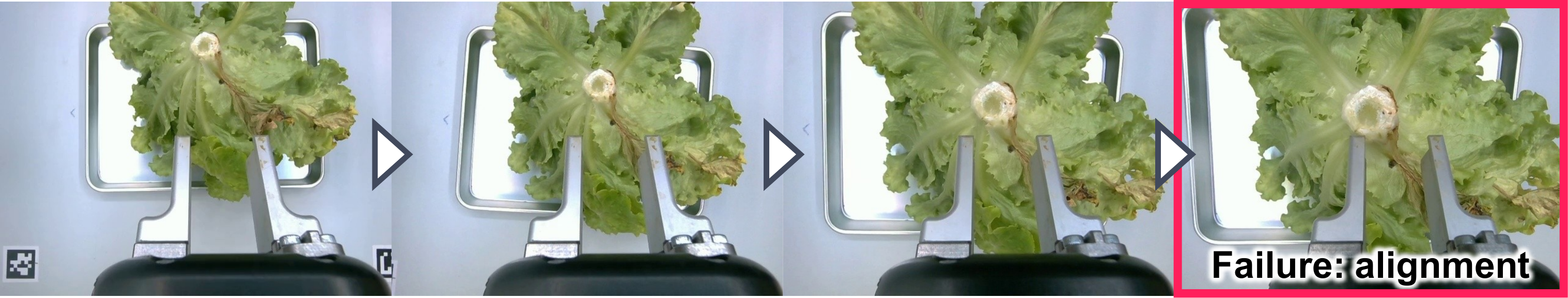}
        \subcaption{DRAIL w/o dual aug.}
        \label{fig:real_lettuce_task_robot_motion_dp}
    \end{minipage}
    \caption{Robot motion by learned policy with each method in the real lettuce defective leaf picking preparation task}
    \label{fig:real_lettuce_task_robot_motion}
\end{figure}

To answer RQ1, we evaluate policies based on task success rates in terms of leaf selection and positional alignment.
Policy motions are shown in \figref{fig:real_lettuce_task_robot_motion}, and task success rates are reported in \tabref{tab:real_lettuce_task_result}.
DRAIL achieves the highest success rates in both leaf selection and positional alignment, while all ablation methods degrade.
This indicates that DRAIL generalizes better to unseen real lettuce appearances.

\begin{table}[t] 
    \centering
    \caption{
    Success rates in the real lettuce defective leaf picking preparation task.
    15 trials were performed.
    }
    \label{tab:real_lettuce_task_result}
    \begin{tabular}{c|cccc}
        \toprule
                  & DRAIL         & DRAIL        & DRAIL            & DRAIL \\
                  &               & w/o task-irr.& w/o task-rel.    & w/o dual  \\
        \midrule                                          
        \midrule                                          
        Leaf      &
        \multirow{2}{*}{\textbf{80\%}} &
        \multirow{2}{*}{67\%} &
        \multirow{2}{*}{53\%} &
        \multirow{2}{*}{47\%} \\[-0.4ex] 
        selection &&&& \\[0.5ex]                           
        Positional  & 
        \multirow{2}{*}{\textbf{73\%}} & 
        \multirow{2}{*}{47\%} & 
        \multirow{2}{*}{40\%} & 
        \multirow{2}{*}{40\%} \\[-0.4ex]
        alignment &&&& \\
        \bottomrule
    \end{tabular}
\end{table}

\begin{figure}[t] 
    \centering
    \includegraphics[width=0.9\linewidth]{figures/colorbar.pdf}
    \begin{minipage}[b]{0.49\linewidth}
        \centering
        \includegraphics[width=0.8\linewidth]{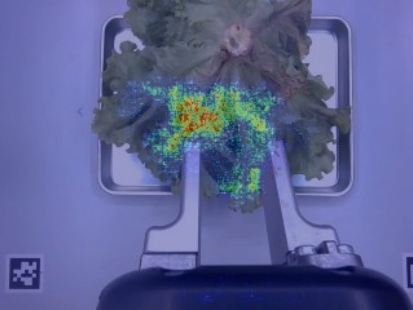}
        \subcaption{DRAIL}
        \label{fig:real_lettuce_task_activation_map_drail}
    \end{minipage}
    \begin{minipage}[b]{0.49\linewidth}
        \centering
        \includegraphics[width=0.8\linewidth]{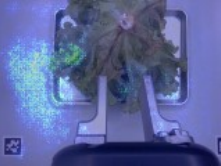}
        \subcaption{DRAIL w/o task-irr. aug.}
        \label{fig:real_lettuce_task_activation_map_rel}
    \end{minipage} 
    \begin{minipage}[b]{0.49\linewidth}
        \centering
        \includegraphics[width=0.8\linewidth]{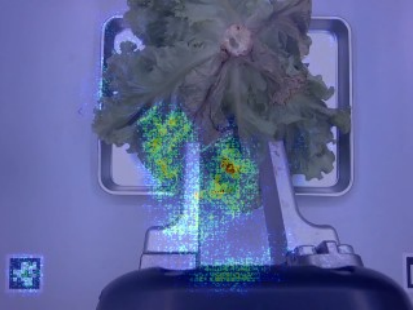}
        \subcaption{DRAIL w/o task-rel. aug.}
        \label{fig:real_lettuce_task_activation_map_ishida}
    \end{minipage} 
    \begin{minipage}[b]{0.49\linewidth}
        \centering
        \includegraphics[width=0.8\linewidth]{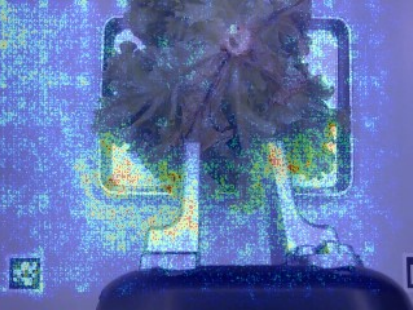}
        \subcaption{DRAIL w/o dual aug.}
        \label{fig:real_lettuce_task_activation_map_dp}
    \end{minipage}
    \begin{minipage}[b]{0.49\linewidth}
        \centering
        \includegraphics[width=0.8\linewidth]{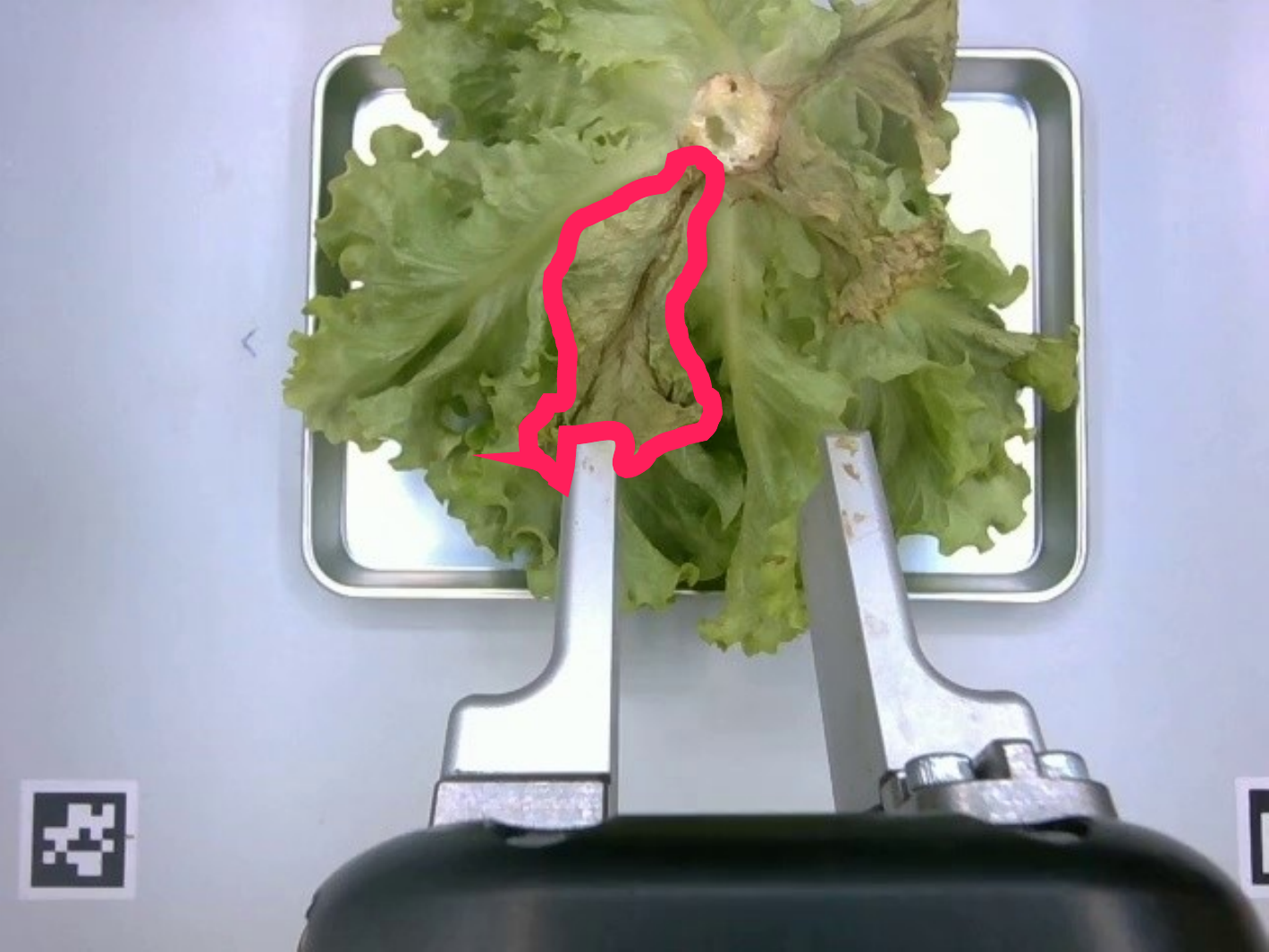}
        \subcaption{Visual observation}
        \label{fig:real_lettuce_task_activation_map_original}
    \end{minipage} 
    \caption{
    Visualization of the attention region of the learned policy by the saliency map in the real lettuce defective leaf picking preparation task.
    In the visual observation shown in (e), the red line highlights the important region that the policy should attend to for the task.
    }
    \label{fig:real_lettuce_task_activation_map}
\end{figure}

To address RQ2, we visualize policy attention regions in the test environment.
Attention regions are shown in \figref{fig:real_lettuce_task_activation_map}.
DRAIL focuses its attention on the base of the most severely defective leaf, which is critical for successful task execution.
In contrast, DRAIL without task-relevant augmentation attends to lettuce leaves; however, it relies on task-irrelevant features on lettuce.
Moreover, both DRAIL without task-relevant augmentation and DRAIL without dual augmentation exhibit dispersed attention across the entire image.
These results suggest that DRAIL suppresses spurious correlations with the background while promoting reliance on task-essential visual features.

\begin{table}[!t] 
    \centering
    \caption{
    Comparison of ARGs of policies learned by each method in the real lettuce defective leaf picking preparation task.
    For each method, ARG was recomputed 20 times with different random samplings of the fixed network.
    Statistical significance was assessed using a paired t-test against DRAIL. 
    ``$\dagger$" indicates a statistically significant difference with $p < 0.05$.
    }
    \label{tab:real_lettuce_task_arg}
    \begin{tabular}{c|cccc}
        \toprule
             & DRAIL                     & DRAIL            & DRAIL                     & DRAIL \\
         ARG &                           & w/o task-irr.    & w/o task-rel.             & w/o dual  \\
        \midrule                                             
        \midrule                                             
        Mean & $\mathbf{8.8\times 10^3}$ & $1.5\times 10^4$ & $9.7\times 10^{4\dagger}$ & $1.7\times 10^{6\dagger}$  \\
        STD  & $\mathbf{5.8\times 10^3}$ & $1.3\times 10^4$ & $1.9\times 10^4$~         & $8.9\times 10^5$~  \\
        \bottomrule
    \end{tabular}
\end{table}

To answer RQ3, we compute ARG as a quantitative measure of generalization to visual variations; results are reported in \tabref{tab:real_lettuce_task_arg}.
DRAIL yields significantly lower ARG values than ablation methods, demonstrating superior visual generalization in the real lettuce defective leaf picking preparation task.
While removing task-relevant augmentation moderately increases ARG, excluding task-relevant or dual augmentation results in a drastic increase.
These findings highlight that task-relevant augmentation plays a critical role in suppressing spurious visual correlations and enables the policy to rely on semantically meaningful visual cues.

Taken together, the above results validated RQ1–RQ3 for this task.
Overall, DRAIL consistently outperforms its ablations in task success, attention visualization, and ARG, demonstrating robust real-world visual generalization enabled by dual-region augmentation.

\section{Discussion}
Future work includes augmentation design for the task-relevant region and data augmentation for multimodal information.
Regarding the augmentation design for the task-relevant region, both task-relevant region extraction and the augmentation design are operated manually based on domain knowledge.
Developing a framework to automatically search for effective augmentation depending on the properties of the task and objects is an important direction.
In this study, we only apply data augmentation to visual observations; however, more advanced agricultural tasks may require leveraging multimodal information such as depth and tactile information.
Further investigation is needed on what kinds of data augmentation should be applied to depth information and other modalities, analogous to RGB images, to promote learning task-essential features.

\section{Conclusion}
This paper proposed DRAIL, a task-relevant and task-irrelevant region-aware augmentation framework for generalizable vision-based imitation learning. DRAIL explicitly separates each visual observation into task-relevant and task-irrelevant regions using domain knowledge and applies differentiated augmentation strategies to each region. Experiments with artificial vegetables and lettuce demonstrated that DRAIL achieved higher task success rates under unseen visual conditions and improved generalization compared to baseline methods. Furthermore, analysis using Absolute RND Gap (ARG) and visualization of visuomotor policy attention quantitatively and qualitatively confirmed that the learned policies relied more on task-essential visual features rather than spurious background correlations.


\end{document}